\documentclass[prodmode,acmcsur]{acmsmall} 

\usepackage{amssymb}
\usepackage{hyperref}
\usepackage{booktabs}
\usepackage{makecell,rotating}
\usepackage{graphicx}
\usepackage{color}
\usepackage{pgf}
\usepackage{dashrule}
\usepackage{tikz}
\usepackage{xcolor}
\usepackage{colortbl}
\usepackage{subfigure}
\usepackage{hyperref}
\usetikzlibrary{arrows, decorations.pathmorphing, backgrounds, positioning, fit, petri, automata}

\usepackage[ruled]{algorithm2e}
\renewcommand{\algorithmcfname}{ALGORITHM}
\SetAlFnt{\small}
\SetAlCapFnt{\small}
\SetAlCapNameFnt{\small}
\SetAlCapHSkip{0pt}
\IncMargin{-\parindent}

\begin{document}

\newlength\savedwidth
\newcommand{\whline}{\noalign{\global\savedwidth\arrayrulewidth
                            \global\arrayrulewidth 1.5pt}%
                   \hline
                   \noalign{\global\arrayrulewidth\savedwidth}}

\markboth{S. Ouyang et al.}{A Survey on Heterogeneous Face Recognition: Sketch, Infra-red, 3D and Low-resolution}

\title{A Survey on Heterogeneous Face Recognition: Sketch, Infra-red, 3D and Low-resolution}
\author{SHUXIN OUYANG
\affil{Beijing University of Posts and Telecommunications}
TIMOTHY HOSPEDALES
\affil{Queen Mary University of London}
YI-ZHE SONG
\affil{Queen Mary University of London}
XUEMING LI
\affil{Beijing University of Posts and Telecommunications}
}

\begin{abstract}

Heterogeneous face recognition (HFR) refers to matching face imagery across different domains. It has received much interest from the research community as a result of its profound implications in law enforcement. A wide variety of new invariant features, cross-modality matching models and heterogeneous datasets being established in recent years. This survey provides a comprehensive review of established techniques and recent developments in HFR. Moreover, we offer a detailed account of datasets and benchmarks commonly used for evaluation. We finish by assessing the state of the field and discussing promising directions for future research.






\end{abstract}





\begin{bottomstuff}

Author's addresses: S. Ouyang, School of Information and Communication Engineering,
Beijing University of Posts and Telecommunications ; T. Hospedales  {and} Y. Song,
Computer Vision Lab, Queen Mary University of London; X. Li, School of Digital Media and Design Art.

\end{bottomstuff}

\maketitle

\section{Introduction}

\label{Introduction}
Face recognition is one of the most studied research topics in computer vision. After over four decades of research, conventional face recognition using visual light under controlled and homogeneous conditions now approaches a mature technology \cite{zhao2003faceRecSurvey}, being deployed at industrial scale for biometric border control \cite{frontex2010biopass} and producing better-than-human performances \cite{SunWT14}.
Much research effort now focuses on uncontrolled, non-visual and heterogeneous face recognition, which remain open questions.
Heterogeneous face recognition (HFR) refers to the problem of matching faces across different visual domains. Instead of working with just photographs, it encompasses the problems of closing the semantic gap among faces captured (i) using different sensory devices (e.g., visual light vs. near-infrared or 3D devices), (ii) under different cameras settings and specifications (e.g., high-resolution vs. low-resolution images), and (iii) manually by an artist and automatically by a digital sensor (e.g., forensic sketches vs. digital photographs).

HFR has grown in importance and interest because heterogeneous sets of facial images must be matched in many practical applications for security and law enforcement as well as multi-media indexing. For example, visual-infrared matching is important for biometric security control, because enrollment images can be taken in controlled a setting with visual light, while probe images may be taken in infra-red if visual lighting in the access control area is not controllable. Meanwhile, sketch-based recognition is important for law-enforcement, where eyewitness sketches should be matched against mugshot databases to identify suspects.

\begin{figure*}[t]
  \centering
    \includegraphics[width=0.85\textwidth]{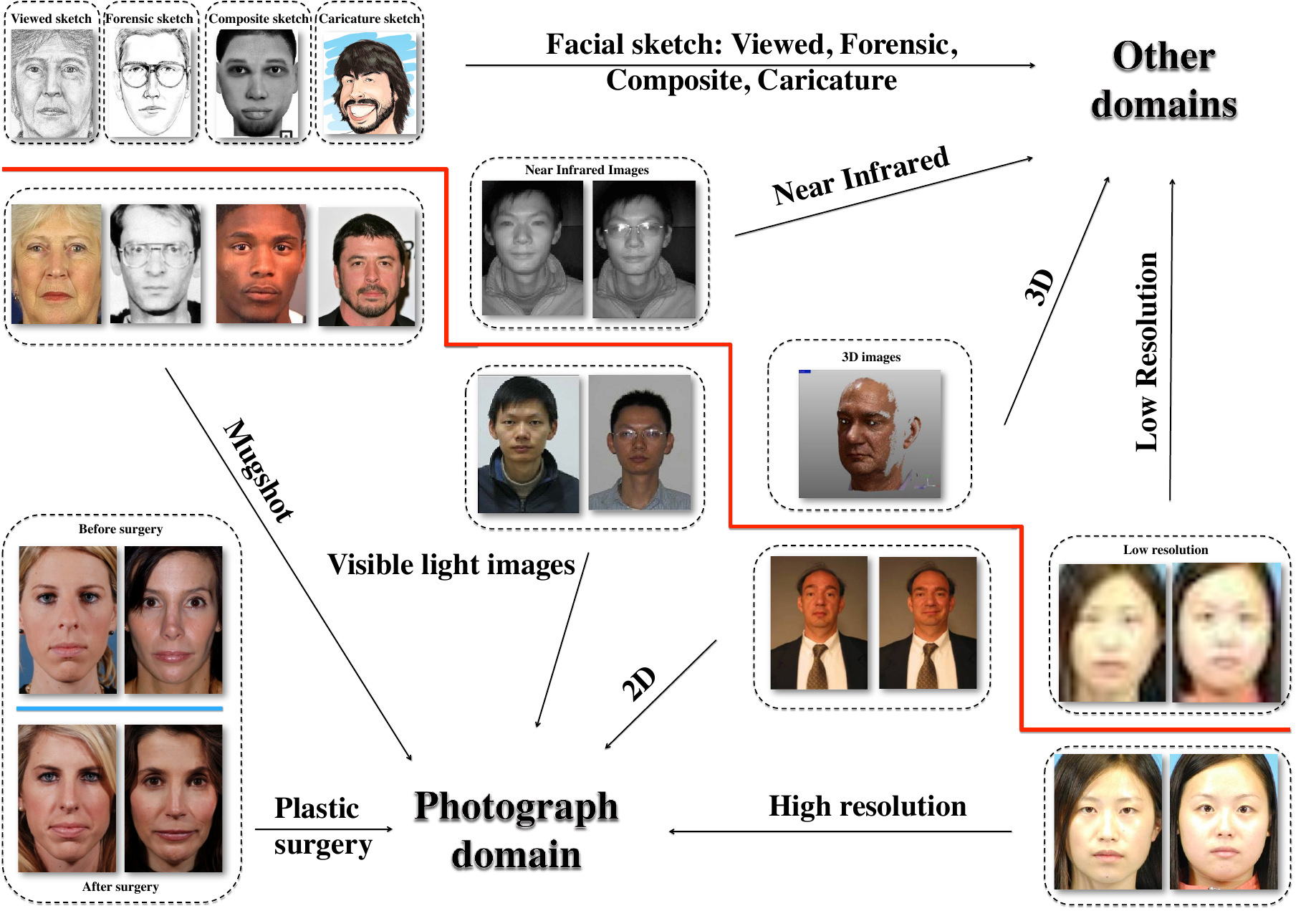}
  \caption{Scope of heterogeneous face recognition studied in this survey.}
  \label{fig:whole}
\end{figure*}

\begin{table}
\tbl{Overview of studies by modality. Superscript $^*$ indicates studies that have been applied to multiple modality pairs.\label{tab:studiesbyModality}}{%
\renewcommand{\arraystretch}{1.5}
\scalebox{0.8}{
\begin{tabular}{ll}
\whline
Domains & Studies\tabularnewline
\hline
 \rowcolor{gray!20}
Sketch-Photo &  \tabularnewline
 \rowcolor{gray!20}
~~~~$\bullet$Viewed & \cite{tang2002facePhotoSketch,wang2009faceSketchRec,galoogahi2012interModalityFace,KianiGaloogahi:2012:FPR:2393347.2396354} \tabularnewline
 \rowcolor{gray!20}
& \cite{6467240,5634507,Klare2010,pramanik2012geo}  \tabularnewline
 \rowcolor{gray!20}
& \cite{Gao2008,4453838,6411670,nejati2011nonArtisticSketch,4244565}  \tabularnewline
 \rowcolor{gray!20}
& \cite{tang2003sketchSynthRecICCV,zhang2011informationTheoretic,huang2013coupledDictionary,1467376}  \tabularnewline
 \rowcolor{gray!20}
& \cite{4217122,Liu:2007:BTI:1625275.1625621,choi2012cvpr,Xiao20091576}$^*$  \tabularnewline
 \rowcolor{gray!20}
& \cite{lin2006interModalityFace,sharma2011pls,huang2013dsr}$^*$  \tabularnewline
 \rowcolor{gray!20}
~~~~$\bullet$Composite & \cite{4244565,Han2013,6330967}  \tabularnewline
 \rowcolor{gray!20}
~~~~$\bullet$Forensic & \cite{bhatt2012memeticSketch,Klare2010,zhang2010sketch,517132,5601735}  \tabularnewline
 \rowcolor{gray!20}
~~~~$\bullet$Caricature & \cite{klare2012charicature}  \tabularnewline
VIS-NIR & \cite{zhu2014transduction,liao2009hfr,yi2007nirVisFaceCCA,chen2009cvpr,goswami2011}\tabularnewline
&\cite{Wang2009,pengfei2012,5597000,lei2012improvedCSR,zhu2013lgh}\tabularnewline
&\cite{huang2013mif,gong2013acpr,5206860,liu2012icb} \tabularnewline
&\cite{lin2006interModalityFace,lei2012improvedCSR,huang2013dsr}$^*$ \tabularnewline
 \rowcolor{gray!20}
2D-3D & \cite{yang2008cca2d3d,Liris-3912,huang2010_2d3d,5539995,rama2006_2d3d,huang2012_2d3d}  \tabularnewline
Low-High & \cite{5206860,5957296,4587810,Zou2010}\tabularnewline
&\cite{jia2005ssr,6117595,li2010lr,Wang2013}\tabularnewline
&\cite{6112780,6467147,5624630,1203152}   \tabularnewline
 &  \cite{5485651,5413920,shekhar2011lrRec,ren2012tip}$^*$  \tabularnewline
&\cite{siena2013btas,ren2011mkl,sharma2011pls,lei2012improvedCSR}$^*$  \tabularnewline
\rowcolor{gray!20}
Plastic surgery & \cite{5492195,6163008,6416547,6327663,Liu:2013aa}\tabularnewline
\whline
\end{tabular}}
}

\end{table}

Nevertheless, HFR poses a variety of serious challenges beyond conventional homogeneous face recognition. These include: (i) comparing single versus multi-channel imagery (e.g., infra-red versus RGB visible light images), (ii) linear and non-linear variations in intensity value due to different specular reflection properties (e.g., infra-red versus RGB), (iii) different coordinate systems (e.g., 2D versus 3D depth images), (iv) reduction of appearance detail (e.g., photo versus sketch, or high versus low-resolution), (v) non-rigid distortion preventing alignment (e.g., photo versus forensic sketch, or comparing imagery before and after plastic surgery). For all these reasons, it is not possible or effective to compare heterogeneous imagery directly as in conventional face recognition.

To address these challenges, the field of HFR has in recent years proposed a wide variety of approaches to bridge the cross-modal gap, thus allowing heterogeneous imagery to be compared for recognition. Research progress in bridging this gap has been assisted by a growing variety of HFR benchmark datasets allowing direct comparison of different methodologies.  This paper provides a comprehensive and up-to-date review of the diverse and growing array of HFR techniques. We categorize them in terms of different modalities they operate across, as well as their strategy used to bridge the cross modal gap -- bringing out some cross-cutting themes that re-occur in different pairs of modalities. Additionally, we summarize the available benchmark datasets in each case, and close by drawing some overall conclusions and making some recommendations for future research.

In most cases HFR involves querying a gallery consisting of high-resolution visible light face photographs using a probe image from an alternative imaging modality. We first break down HFR research in the most obvious way by the pairs of imagery considered. We consider four cross-modality applications: sketch-based, infra-red based, 3D-based and high-low resolution matching; as well as one within-modality application of post-surgery face matching. More specifically they are:

\begin{itemize}
\item \textbf{Sketch}: Sketch-based queries are drawn or created by humans rather than captured by an automatic imaging device. The major example application is facial sketches made by law enforcement personal based on eye-witness description. The task can be further categorized into four variants based on level of sketch abstraction, as shown in the left of Fig~\ref{fig:whole}. 

\item \textbf{Near Infrared}: Near Infrared (NIR) images are captured by infrared rather than visual-light devices. NIR capture may be used to establish controlled lighting conditions in environment where visual light is not controllable. The HFR challenge comes in matching NIR probe images against visual light images. A major HFR application is access control, where enrollment images may use visual light, but access gates may use infra-red.

\item\textbf{3D}: Another common access control scenario relies on an enrollment gallery of 3D images and 2D probe images. As the gallery images contain more information than the probe images, this can potentially outperform vanilla 2D-2D matching, if the heterogeneity problem can be solved.


\item \textbf{Low-Resolution:} Matching low-resolution  against  high-resolution images is a topical challenge under contemporary security considerations. A typical scenario is that a high-resolution `watch list' gallery is provided, and low-resolution facial images taken at standoff distance by surveillance cameras  are used as probes. 


\item \textbf{Within-modality Heterogeneity}: Various within-modality effects can also transform facial images significantly enough to seriously challenge conventional recognition systems. In this survey we also discuss the recently topical area of recognition across plastic surgery, that is also in demand due to implications for security and forensics. 

\end{itemize}

Fig~\ref{fig:whole} offers an illustrative summary of the five categories of HFR literature covered in this survey. Tab~\ref{tab:studiesbyModality} further summarizes all the studies reviewed broken down by the modality or modalities considered.

Related areas that are not covered by this review include (homogeneous) 3D \cite{bowyer2006faceRec2d3d} and infra-red \cite{kong2005irVisualFaceReview} matching,  fusing multiple modalities \cite{bowyer2006faceRec2d3d,kong2005irVisualFaceReview}. View \cite{zhang2009faceRecAcrossPose} and illumination \cite{zou2007illuminationInvarFaceRec} invariant recognition are also related in that there exists a strong covariate between probe and gallery images, however we do not include these as good surveys already exist. A good survey about face-synthesis \cite{wang2014faceHallucination} is more relevant to this work, however we consider the broader problem of cross-domain matching.

Most HFR studies focus their contribution on improved methodology to bridge the cross-modal gap, thus allowing conventional face recognition strategies to be used for matching. Even across the wide variety of application domains considered above, these methods can be broadly categorized into three groups of approaches: (i) those that synthesize one modality from another, thus allowing them to be directly compared; (ii) those that engineer or learn feature representations that are variant to person identity while being more invariant to imaging modality than raw pixels; and (iii) those that project both views into a common space where they are more directly comparable. We will discuss these in more detail in later sections.

The main contributions of this paper are summarized as follows:
\begin{enumerate}
  \item We perform an up-to-date survey of HFR literature
  \item We summarize all common public HFR datasets introduced thus far
  \item We extract some cross-cutting themes face recognition with a cross-modal gap
  \item We draw some conclusions about the field, and offer some recommendations about future work on HFR
\end{enumerate}

The rest of this paper is organized as follow: In Section~\ref{sec:system}, we provide an overview of a HFR system pipeline, and highlight some cross-cutting design considerations. In Section~\ref{sec:sketch}, we provide a detailed review of methods for matching facial sketches to photos and a systematic introduction of the most widely used facial sketches datasets. In Section~\ref{nir}, we describe approaches for matching near-infrared to visible light face images in detail. In Section~\ref{sec:2d3d}, we focus on matching 2D probe images against a 3D enrollment gallery. Section~\ref{sec:LR_HR} discusses methods for matching low-resolution face images to high-resolution face images. Finally, Section~\ref{within} discusses matching faces across plastic surgery variations. We conclude with a discussion of current issues and recommendations about future work on HFR.
\section{Outline of a HFR system}\label{sec:system}

\label{sketch}

\begin{figure}[t]
\centering
\includegraphics[height = 2.3in]{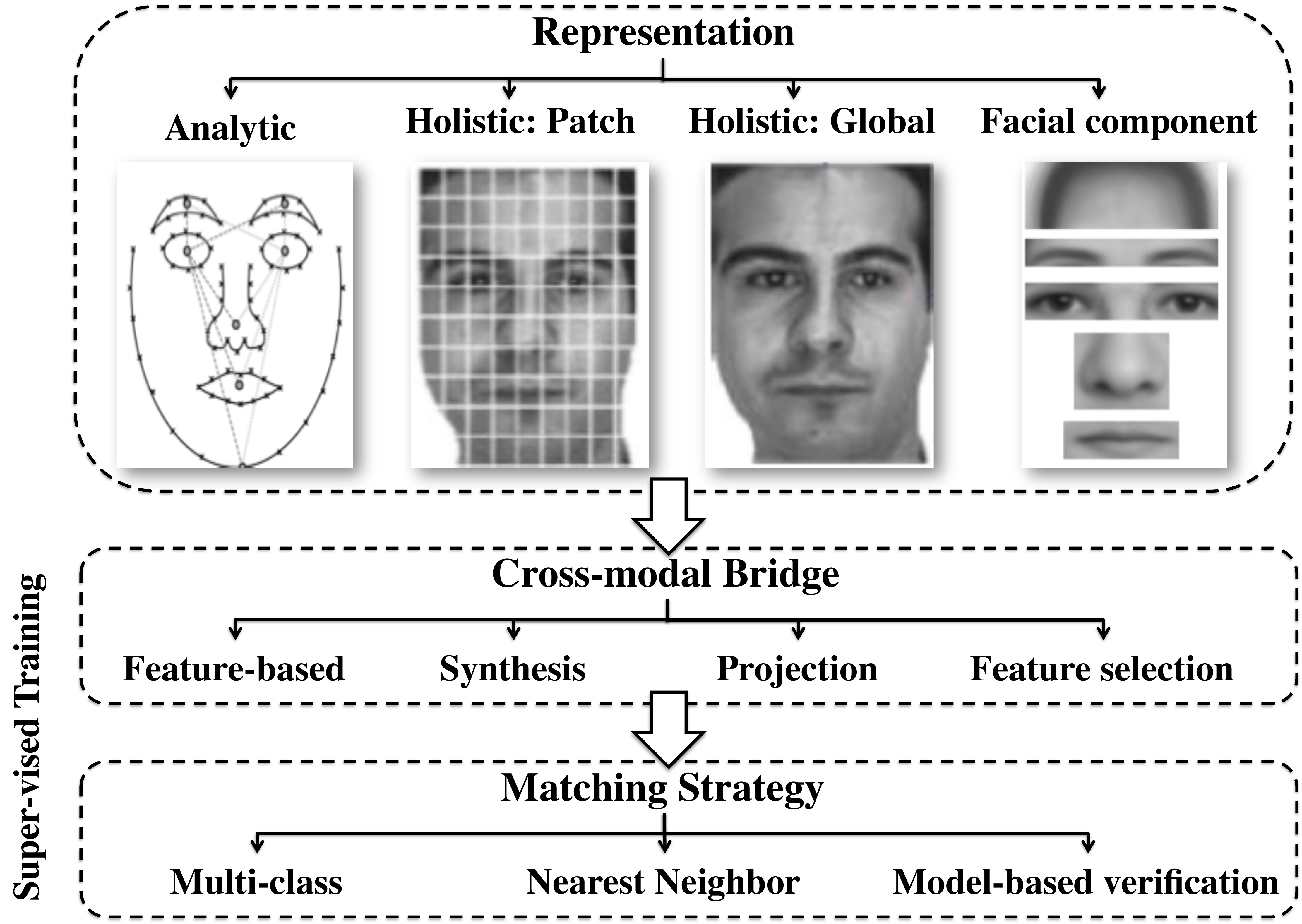}
   \caption{Overview of an abstract HFR pipeline.}
\label{fig:overview}
\end{figure}

In this section, we present an abstract overview of a HFR pipeline, outlining the key steps  and the main types of  strategies available at each stage.
A HFR system can be broken into three major components, each corresponding to an important design decision: representation, cross-modal strategy and matching strategy (Fig~\ref{fig:overview}). Of these components, the first and third have analogues in homogeneous face recognition, while the cross-modal bridge strategy is unique to HFR. Accompanying Fig~\ref{fig:overview}, Tab~\ref{tab:overview} breaks down all the papers reviewed in this survey by their choices about these design decisions.

\subsection{Representation}

The first component of a HFR system determines how the face image in each modality is represented. Common options for representations (Fig~\ref{fig:overview}, top) include analytic, component-based, patch-based, and holistic.

\noindent\textbf{Analytic representations}\quad \cite{nejati2011nonArtisticSketch,4244565,pramanik2012geo} detect facial components and fiducial points, allowing the face to be modeled geometrically, e.g., using point distribution models \cite{nejati2011nonArtisticSketch,4244565}. This representation has the advantage that if a model can be fit to a face in each modality, then the analytic/geometric representation is relatively invariant to modality, and to precise alignment of the facial images. However, it is not robust to errors in face model fitting and may require manual intervention to avoid this \cite{4244565}. Moreover geometry is not robust to facial expression  \cite{pramanik2012geo}, and does not exploit texture information by default. 

\noindent\textbf{Component-based representations}\quad detect face parts (e.g., eyes and mouth), and represents the appearance of each individually \cite{liu2012icb,Han2013}. This allows the informativeness of each component in matching to be measured separately \cite{liu2012icb}; and if components can be correctly detected and matched it also provides some robustness to both linear and non-linear misalignment across modalities \cite{Han2013}. However, a component-fusion scheme is then required to produce an overall match score.

\noindent\textbf{Global holistic representations}\quad represent the whole face image in each modality with a single vector \cite{yi2007nirVisFaceCCA,tang2002facePhotoSketch,lin2006interModalityFace}. Compared to analytic and component-based approaches, this has the advantage of encoding all available appearance information. However, it is sensitive to alignment and expression/pose variation, and may provide a high-dimensional feature vector that risks over-fitting \cite{tan2006singleImageFaceRec}.

\noindent \textbf{Patch-based holistic representations}\quad encode the appearance of each image in patches with a feature vector per patch \cite{wang2009faceSketchRec,1467376,galoogahi2012interModalityFace,KianiGaloogahi:2012:FPR:2393347.2396354,bhatt2012memeticSketch}. Subsequent strategies for using the patches vary, including for example concatenation into a very large feature vector \cite{Klare2010} (making it in effect a holistic representation), or learning a mapping/classifier per patch \cite{zhang2011informationTheoretic}. The latter strategy can provide some robustness if the true mapping is not constant over the whole face, but does require a patch fusion scheme.


\begin{table*}
\tbl{\label{tab:overview}Overview of heterogeneous face recognition steps and typical strategies for each.}{
\renewcommand{\arraystretch}{1.3}
\scalebox{0.8}{
\begin{tabular}{lrl}
\whline
Component & Approach & Representative Examples\tabularnewline
\hline
\rowcolor{gray!20}
Representation & Analytic & Active Shape \& Point Distribution Models \cite{nejati2011nonArtisticSketch,4244565}\tabularnewline
\rowcolor{gray!20}
 & & Relative Geometry \cite{pramanik2012geo} \tabularnewline
\rowcolor{gray!20}
 & Global Holistic & Whole image \cite{yi2007nirVisFaceCCA,tang2002facePhotoSketch,tang2003sketchSynthRecICCV}\tabularnewline
 \rowcolor{gray!20}
 &  & Whole image\cite{liao2009hfr,Wang2009,5206860,lei2012improvedCSR}\tabularnewline
  \rowcolor{gray!20}
 &  & Whole image\cite{sharma2011pls,zhu2013lgh,Li2009ACML,siena2013btas}\tabularnewline
 \rowcolor{gray!20}
 & Global Patch & Regular grid of patches \cite{wang2009faceSketchRec,1467376,galoogahi2012interModalityFace}\tabularnewline
\rowcolor{gray!20}
 &  & Regular grid of patches \cite{KianiGaloogahi:2012:FPR:2393347.2396354,Klare2010,bhatt2012memeticSketch}\tabularnewline
\rowcolor{gray!20}
 &  & Regular grid of patches \cite{6411670,yang2008cca2d3d,chen2009cvpr}\tabularnewline
 \rowcolor{gray!20}
 & Facial Component & Active Shape Model Detection \cite{Han2013} \tabularnewline
 \rowcolor{gray!20}
 &  & Rectangular patches \cite{liu2012icb} \tabularnewline
Cross domain & Feature-based & LBP \cite{bhatt2012memeticSketch,liao2009hfr,chen2009cvpr,Han2013}\tabularnewline
 & & Gabor \cite{KianiGaloogahi:2012:FPR:2393347.2396354},
SIFT \cite{Klare2010}, SSIM \cite{6411670}\tabularnewline
 & & CITE \cite{zhang2011informationTheoretic}, LGH \cite{zhu2013lgh}, HOAG \cite{galoogahi2012interModalityFace}\tabularnewline
 & Projection & CDFE \cite{lin2006interModalityFace}, Common Basis \cite{Klare2010} \tabularnewline
 &  & CSR \cite{5206860,lei2012improvedCSR,liu2012icb} RS-LDA \cite{wang2009faceSketchRec} \tabularnewline
 &  & CCA \cite{yi2007nirVisFaceCCA,yang2008cca2d3d,Li2009ACML},  PLS \cite{sharma2011pls}
  \tabularnewline
 &  & Adaboost \cite{liu2012icb,liao2009hfr}, Max-margin \cite{siena2013btas} \tabularnewline
  &  & Kernel LDA \cite{1467376}, Sparse Coding \cite{huang2013coupledDictionary,shekhar2011lrRec}\tabularnewline
 & Synthesis & NN \cite{Wang2009,chen2009cvpr}, MRF \cite{wang2009faceSketchRec} \tabularnewline
  &  &  Eigentransform \cite{tang2002facePhotoSketch,tang2003sketchSynthRecICCV}  \tabularnewline
 & & LLE \cite{1467376}, Relationship learning \cite{5957296}\tabularnewline
 \rowcolor{gray!20}
Matching & Multi-class  & NN \cite{tang2002facePhotoSketch,Klare2010,tang2003sketchSynthRecICCV,chen2009cvpr}\tabularnewline
 \rowcolor{gray!20}
 &   & NN \cite{Li2009ACML,6411670,nejati2011nonArtisticSketch,5206860}\tabularnewline
  \rowcolor{gray!20}
 &   & NN \cite{lei2012improvedCSR,sharma2011pls,zhu2013lgh}\tabularnewline
 \rowcolor{gray!20}
 &  & NN with $\chi^{2}$ \cite{galoogahi2012interModalityFace,KianiGaloogahi:2012:FPR:2393347.2396354,liu2012icb} \tabularnewline
 \rowcolor{gray!20}
 &  & NN with HI \cite{Han2013},NN with Cosine \cite{yi2007nirVisFaceCCA,yang2008cca2d3d}\tabularnewline
 \rowcolor{gray!20}
 & Multi-class (Tr) & Bayesian \cite{tang2003sketchSynthRecICCV}, MA metric learning for
$\chi^{2}$ NN \cite{bhatt2012memeticSketch}\tabularnewline
 \rowcolor{gray!20}
 &  & SVM \cite{5957296}\tabularnewline
 \rowcolor{gray!20}
 & Verification (Tr) & Similarity threshold (Cosine) \cite{liao2009hfr}, SVM \cite{klare2012charicature}\tabularnewline
  \rowcolor{gray!20}
 & & Logistic Regression \cite{klare2012charicature}\tabularnewline
\whline
\end{tabular}
}
}

\end{table*}

\subsection{Cross-modal bridge strategy}
The key HFR challenge of cross-modality heterogeneity typically necessitates an explicit strategy to deal with the cross-modal gap.  This component uniquely distinguishes HFR systems from conventional within-modality face recognition. Most HFR studies focus their effort on developing improved strategies for this step. Common strategies broadly fall into the categories: feature design, cross-modal synthesis and subspace projection. These strategies are not exclusive, and many studies employ or contribute to more than one \cite{Klare2010,wang2009faceSketchRec}.

\noindent\textbf{Feature design}\quad strategies \cite{galoogahi2012interModalityFace,KianiGaloogahi:2012:FPR:2393347.2396354,Klare2010,bhatt2012memeticSketch} focus on engineering or learning features that are invariant to the modalities in question, while simultaneously being discriminative for person identity. Typical strategies include variants on SIFT \cite{Klare2010} and LBP \cite{bhatt2012memeticSketch}.

\noindent\textbf{Synthesis}\quad approaches focus on synthesizing one modality based on the other \cite{tang2002facePhotoSketch,wang2009faceSketchRec}. Typical methods include eigentransforms \cite{tang2002facePhotoSketch,tang2003sketchSynthRecICCV}, MRFs \cite{wang2009faceSketchRec}, and LLE \cite{1467376}. The synthesized image can then be used directly for homogeneous matching. Of course, matching performance is critically dependent on the fidelity and robustness of the synthesis method.

\noindent\textbf{Projection}\quad approaches aim to project both modalities of face images to a common subspace in which they are more comparable than in the original representations \cite{lin2006interModalityFace,Klare2010,yi2007nirVisFaceCCA}. Typical methods include linear discriminant analysis (LDA) \cite{wang2009faceSketchRec}, canonical components analysis (CCA) \cite{yi2007nirVisFaceCCA,yang2008cca2d3d}, partial least squares (PLS) and common basis \cite{Klare2010} encoding.

A noteworthy special case of projection-based strategies is those approaches that perform \emph{feature selection}. Rather than mapping all input dimensions to a subspace, these approaches simply discover which subset of input dimensions are the most useful (modality invariant) to compare across domains, and ignore the others \cite{liu2012icb,liao2009hfr}, for example using Adaboost.
 
\subsection{Matching strategy}

Once an effective representation has been chosen, and the best effort made to bridge the cross-modal heterogeneity, the final component of a HFR system is the matching strategy. Matching-strategies may be broadly categorized as multi-class classifiers (one class corresponding to each identity in the gallery), or model-based verifiers.

\noindent\textbf{Multi-class classifiers}\quad pose the HFR task as a multi-class-classification problem. The probe image (after the cross-modal transform in the previous section) is classified into one of the gallery classes/identities. Typically simple classifiers are preferred because there are often only one or a few gallery image(s) per identity, which is too sparse to learn sophisticated classifiers. Thus \emph{Nearest-Neighbor (NN)} \cite{tang2002facePhotoSketch,lin2006interModalityFace,Klare2010,yi2007nirVisFaceCCA} is most commonly used to match against the gallery \cite{tang2002facePhotoSketch}. NN classifiers can be defined with various distance metrics, and many studies found $\chi^2$ \cite{galoogahi2012interModalityFace,KianiGaloogahi:2012:FPR:2393347.2396354} or cosine \cite{yang2008cca2d3d} to be most effective than vanilla euclidean distance. An advantage of NN-based approaches is that they do not require an explicit training step or training data. However, they can be enhanced with metric-learning \cite{bhatt2012memeticSketch} if annotated cross-domain training data is available.
 
\noindent\textbf{Model-based verification strategies}\quad pose HFR as a binary, rather than multi-class, classification problem \cite{liao2009hfr,klare2012charicature}. These take a pair of heterogeneous images as input, and output one or zero according to if they are estimated to be the same person. An advantage of verification over classification strategies is robustness and data sufficiency. In many scenarios there is only one cross-modal face pair per person. Thus classification strategies have one instance per class (person), and risk over fitting. In contrast, by transforming the problem into a binary one, all true pairs of faces form the positive class and all false pairs form the negative class, resulting in a much larger training set, and hence a stronger and more robust classifier.

We note that some methodologies can be interpreted as either cross-domain mappings or matching strategies. For example, some papers \cite{wang2009faceSketchRec} present LDA as a recognition mechanism. However, as it finds a projection that maps images of one class (person identity) closer together, it also has a role in bridging the cross-modal gap when those images are heterogeneous. Therefore for consistency, we categorize LDA and the like as cross-domain methods.

\subsection{Formalizations}

Many HFR methods can be seen as special cases of a general formalization given in Eq.~\ref{eq:genMatch}. Images in two modalities $\mathbf{x}^a$ and $\mathbf{x}^b$ are input; non-linear feature extraction $F$ may be performed; and some matching function $M$ then compares the extracted features; possibly after taking linear transforms $W^a$ and $W^b$ of each feature. 
\begin{eqnarray}
 &  & M\left(W^{a}F(\mathbf{x}_{i}^{a}),W^{b}F(\mathbf{x}_{j}^{b})\right).\label{eq:genMatch}
 \end{eqnarray}
\noindent many studies reviewed in this paper can be seen as providing different strategies for determining the mappings $W^a$ and $W^b$ or parameterizing functions $M$ and $F$.

\noindent\textbf{Matching Strategies}\quad Many matching strategies can be seen as design decisions about $M(\cdot,\cdot)$. For example, in the case of NN matching, the closest match $j^*$ to a probe $i$ is returned. Thus $M$ defines the distance metric $\left\Vert\cdot\right\Vert$, as in Eq.~(\ref{eq:matchNN}). In the case of model based verification strategies, a match between $i$ and $j$ may be declared depending on the outcome of a model's (e.g., Logistic Regression \cite{klare2012charicature},  SVM \cite{klare2012charicature}) evaluation of the two projections (e.g., their difference), e.g., Eq.~(\ref{eq:matchVerif}). In this case, matching methods  propose different strategies to determine the parameters $\mathbf{w}$ of the decision function.
\begin{eqnarray}
j^{*} & = & \arg\min_{j}\left\Vert W^{a}F(\mathbf{x}_{i}^{a})-W^{b}F(\mathbf{x}_{j}^{b})\right\Vert \label{eq:matchNN}\\
\mbox{match} & \mbox{iff} & \mathbf{w}^{T}\left|W^{a}F(\mathbf{x}_{i}^{a})-W^{b}F(\mathbf{x}_{j}^{b})\right|>0\label{eq:matchVerif}
\end{eqnarray}

\noindent\textbf{Cross-domain Strategies}\quad Feature-centric cross-domain strategies \cite{galoogahi2012interModalityFace,KianiGaloogahi:2012:FPR:2393347.2396354,Klare2010,bhatt2012memeticSketch,liao2009hfr,chen2009cvpr,Han2013,6411670,zhu2013lgh,zhang2011informationTheoretic} can be seen as designing improved feature extractors $F$. While projection/synthesis strategies can be seen as different approaches to finding the projections $W^a$ and $W^b$ to help make the domains more comparable. For example synthesis strategies \cite{Wang2009,5957296} may set $W^a=I$, and search for the projection $W^b$ so that $\left|F(\mathbf{x}^a_i)-W^b F(\mathbf{x}^a_i)\right|$ is minimized. CCA \cite{yi2007nirVisFaceCCA,yang2008cca2d3d} strategies search for $W^a$ and $W^b$ such that $\left|W^aF(\mathbf{x}^a_i)-W^b F(\mathbf{x}^a_i)\right|$ is minimized for cross-modal pairs of the same person $i$. While LDA \cite{wang2009faceSketchRec} strategies search for a single projection $W$ such that $\left|WF(\mathbf{x}^a_i)-W F(\mathbf{x}^a_j)\right|$ is minimized when $i=j$ and maximized when $i\neq j$.

\subsection{Summary and Conclusions}

HFR methods explicitly or implicitly make design decisions about three stages of representation, cross-domain mapping and matching (Fig~\ref{tab:overview}). An important factor in the strengths and weaknesses of each approach arises from the use of supervised training in either or both of the latter two stages (Fig~\ref{fig:overview}).  

\noindent\textbf{Use of training data}\quad An important property of HFR systems is whether annotated cross-modal training data is required/exploited. This has practical consequences about whether an approach can be applied in a particular application, and its expected performance.
Since a large dataset of annotated cross-modal pairs may not be available, methods that require no training data (most feature-engineering and NN matching approaches \cite{galoogahi2012interModalityFace,KianiGaloogahi:2012:FPR:2393347.2396354,6411670,Han2013}) are advantageous.

On the other hand, exploiting available annotation provides a critical advantage to learn better cross-domain mappings, and many discriminative matching approaches. Methods differ in how strongly they exploit available supervision. For example CCA tries to find the subspace where cross-modal pairs are most similar \cite{yi2007nirVisFaceCCA,yang2008cca2d3d}. In contrast, LDA simultaneously finds a space where cross-modal paris are similar and also where different identities are well separated \cite{wang2009faceSketchRec}, which more directly optimizes the desired outcome of high cross-modal matching accuracy.

\noindent\textbf{Heterogeneous Feature Spaces}\quad A second important model-dependent property is whether the model can deal with heterogeneous data dimensions. In some cross-modal contexts (photo-sketch, VIS-NIR), while the data distribution is heterogeneous, the data dimensions can be the same; while in 2D-3D or low-high, the data dimensionality may be fundamentally different. In the latter case approaches that require homogeneous dimensions such as LDA may not be applicable, while others such as CCA and PLS can still apply.

\section{Matching facial sketches to images}\label{sec:sketch}

The problem of matching facial sketches to photos is commonly known as sketch-based face recognition (SBFR). It typically involves a gallery dataset  of visible light images and a probe dataset of facial sketches. An important application of SBFR is assisting law enforcement to identify suspects by retrieving their photos automatically from existing police databases. Over the past decades, it has been accepted as an effective tool in law reinforcement. In most cases, actual photos of suspects are not available, only sketch drawings based on the recollection of eyewitnesses. The ability to match forensic sketches to mug shots not only has the obvious benefit of identifying suspects, but moreover allows the witness and artist to interactively refine the sketches based on similar photos  retrieved \cite{wang2009faceSketchRec}.


SBFR datasets can be categorized based on how the sketches are generated, as shown in Fig~\ref{fig:facial-sketch-mugshot}: (i) viewed sketches, where artists are given mugshots as reference, (ii) forensic sketches, where sketches are hand-drawn by professional artists based on recollections of witnesses, (iii) composite sketches, where rather than hand-drawn they were produced using specific software, and (iv) caricature sketches, where facial features are exaggerated.

The majority of existing SBFR studies focused on recognizing viewed hand drawn sketches. This is not a realistic use case -- a sketch would not be required if a photo of a suspect is readily available. Yet studying them is a middle ground toward understanding forensic sketches -- viewed sketch performance should reflect the ideal forensic sketch performance when all details are remembered and communicated correctly. Research can then focus on making good viewed sketch methods robust to lower-quality forensic sketches.



%

\subsection{Categorization of facial sketches}

\begin{figure}[t]
\centering
  \includegraphics[height = 2in]{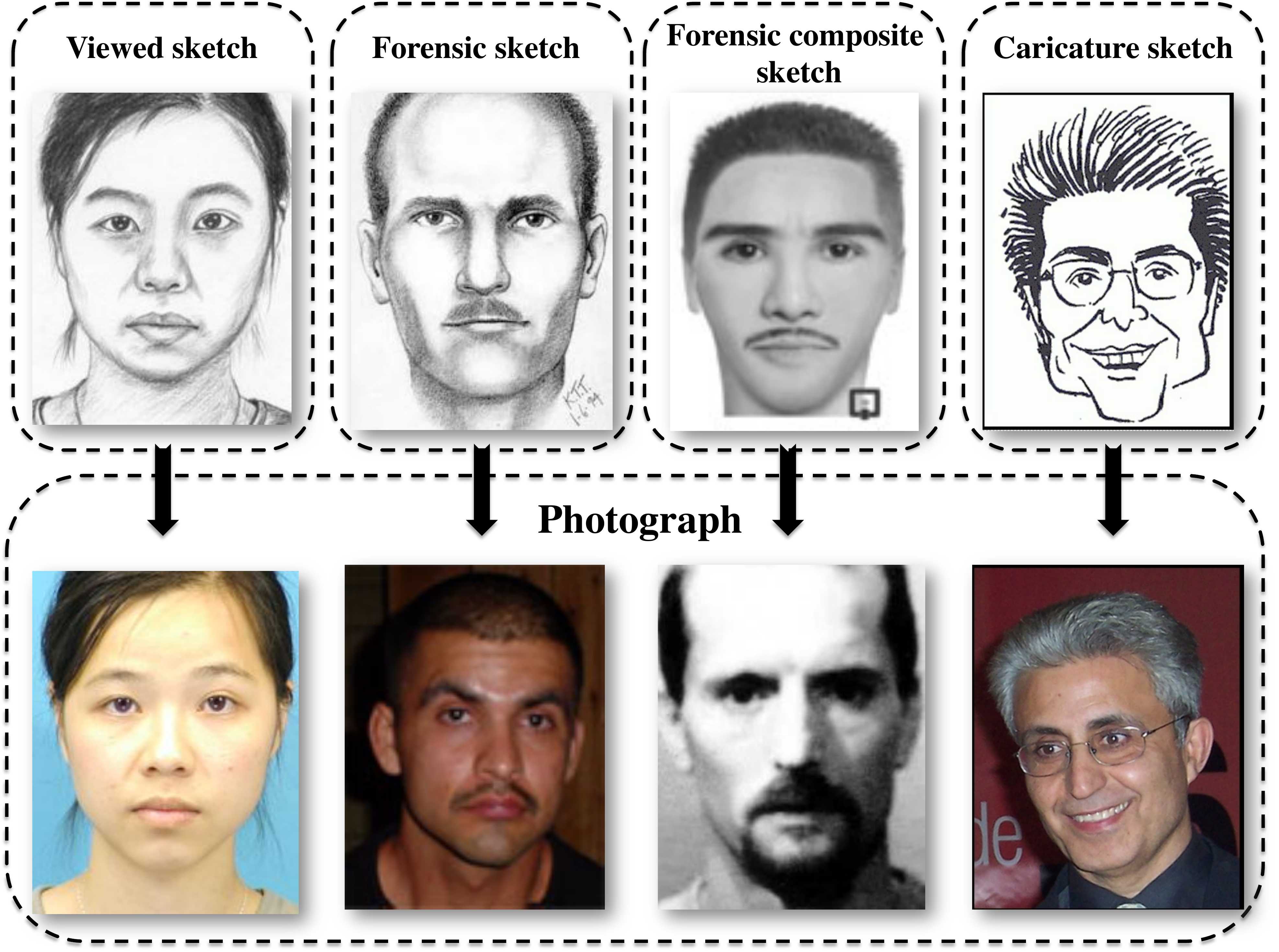}
   \caption{Facial sketches and corresponding mugshots: viewed sketch, forensic hand drawn sketch, forensic composite sketch, caricature sketch and their corresponding facial images}
\label{fig:facial-sketch-mugshot}
\end{figure}

Facial sketches can be created either by an artist or by software, and are referred to as \emph{hand-drawn} and \emph{composite} respectively. Meanwhile depending on whether the artist observes the actual face before sketching, they can also be categorized as \emph{viewed} and \emph{forensic} (unviewed). Based on these factors, we identify four typically studied categories of facial sketches:

\begin{figure}
\centering
\begin{minipage}[t]{0.34\linewidth}
\centering
\includegraphics[height=1.1in]{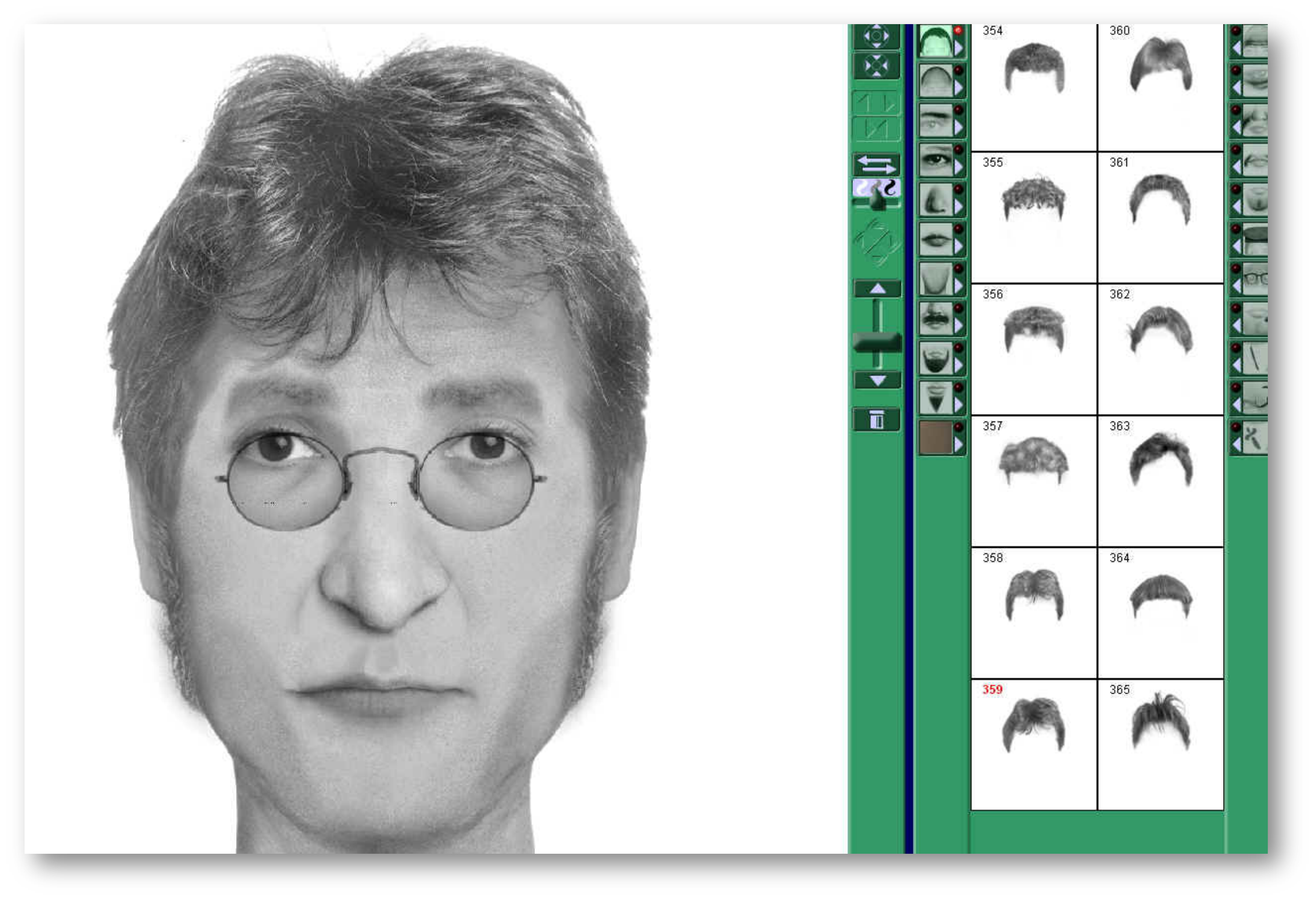}
\centerline{\scriptsize{(a)FACES 4.0~\cite{face2011}}}
\end{minipage}%
\begin{minipage}[t]{0.32\linewidth}
\centering
\includegraphics[height=1.1in]{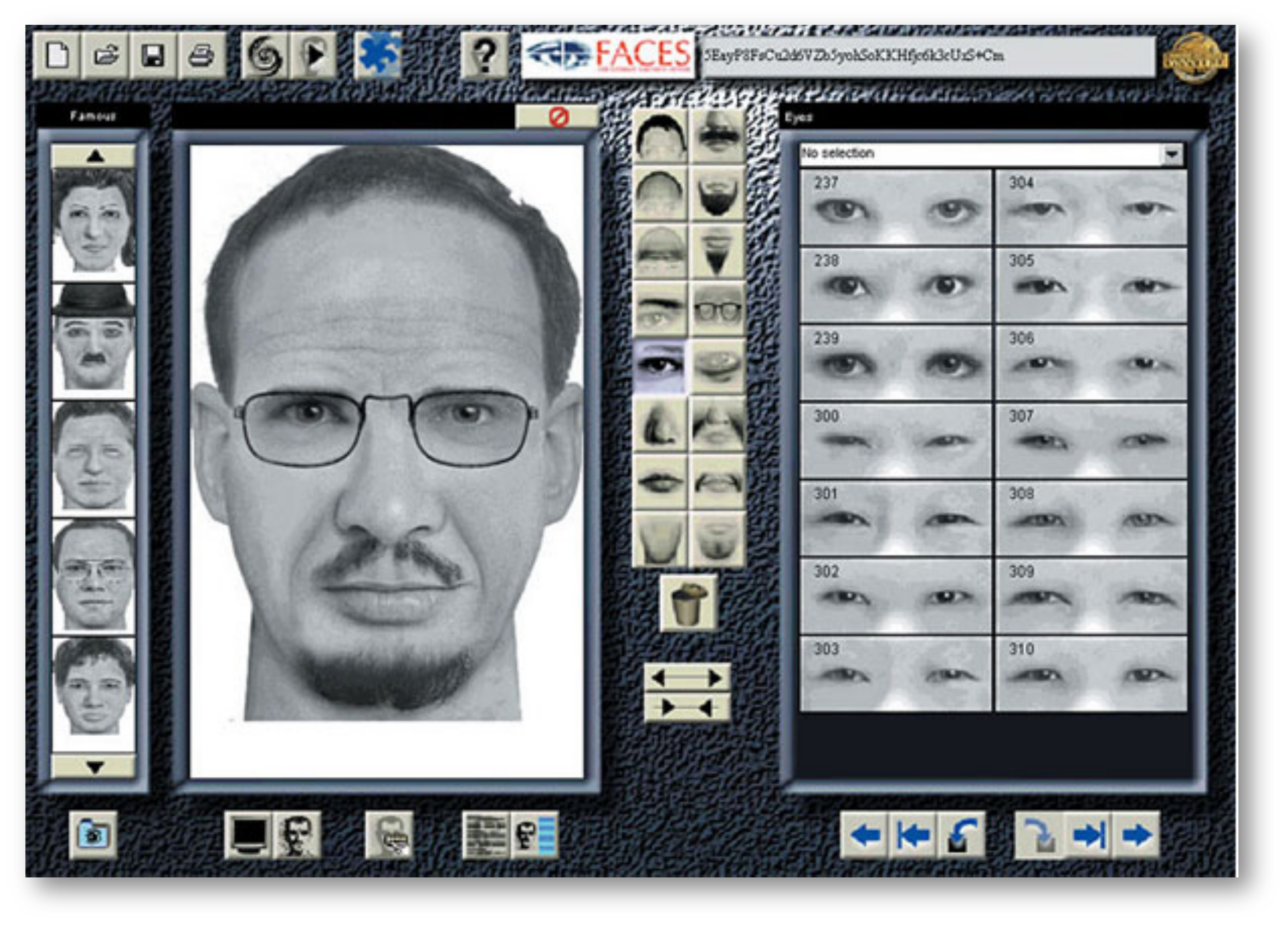}
\centerline{\scriptsize{(b)FACES~\cite{face2011}}}
\end{minipage}
\begin{minipage}[t]{0.32\linewidth}
\centering
\includegraphics[height=1.1in]{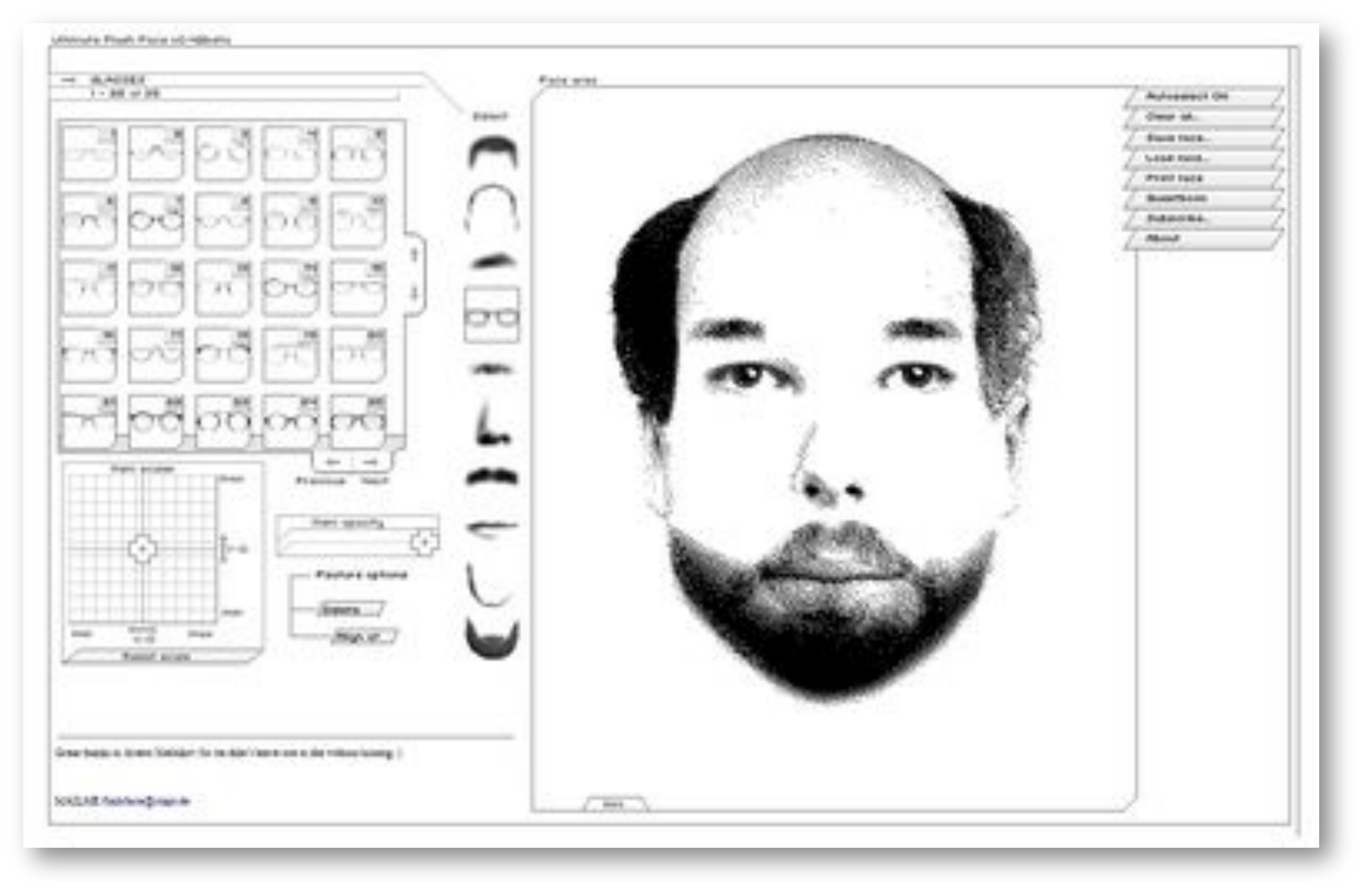}
\centerline{\scriptsize{(c)IdentiKit~\cite{identi}}}
\end{minipage}
\caption{Examples of different kind of composite sketch softwares}
  \label{fig:composite-sketch-software}
\end{figure}

\begin{itemize}
\item \textbf{Forensic hand drawn sketches}: These are produced by a forensic artist based on the description of a witness (\cite{6612993}), as illustrated in the second column of Fig~\ref{fig:facial-sketch-mugshot}. They have been used by police since the 19th century, however they have been less well studied by the recognition community.

\item \textbf{Forensic composite sketches}: They are created by computer software (Fig~\ref{fig:composite-sketch-software}) with which a trained operator selects various facial components  based on the description provided by a witness. An example of a resulting composite sketch is shown in the third column of Fig~\ref{fig:facial-sketch-mugshot}. It is reported that  80\% of law enforcement agencies use some form of software to create facial sketches of suspects \cite{254904}. The most widely used software for generating facial composite sketches are IdentiKit~\cite{identi}, Photo-Fit~\cite{Wells2007}, FACES~\cite{face2011}, Mac-a-Mug~\cite{Wells2007}, and EvoFIT~\cite{FrowdHC04}. It is worth nothing that due to the limitations of such software packages, less facial detail can be presented in composite sketches compared with hand-drawn sketches.

\item \textbf{Viewed hand drawn sketches}: In contrast to forensic sketches that are unviewed, these are sketches drawn by artists by while looking at a corresponding photo, as illustrated in the first column of Fig~\ref{fig:facial-sketch-mugshot}. As such, they are the most similar to the actual photo.

\item \textbf{Caricature}: In contrast to the previous three categories, where the goal is to render the face as accurately as possible, caricature sketches are purposefully dramatically exaggerated. This adds a layer of abstractness that makes their recognition by conventional systems much more difficult. See fourth column of Fig~\ref{fig:facial-sketch-mugshot} for an example.  However,  they are interesting to study because they allow the robustness of SBFR systems to be rigorously tested, and because there is evidence that humans remember faces in a caricatured form, and can recognize them even better than accurate sketches \cite{nejati2011nonArtisticSketch,zhang2010sketch,turk1991eigenfaces}.
\end{itemize}

\subsection{Facial sketch datasets}


\begin{table*}
\tbl{Existing facial sketch benchmark datasets. \label{Tab:existing-dataset}}{%
\renewcommand{\arraystretch}{1.5}
\scalebox{0.7}{
\begin{tabular}{lcccc}
\whline
Datasets & Pairs of Sketch/Photo & Viewed or Forensic & Composite or Hand drawn & Availability \tabularnewline
\hline
\rowcolor{gray!20}
CUFS \cite{wang2009faceSketchRec} &  606 & Viewed & Hand drawn & CUHK: Free to  download \tabularnewline
\rowcolor{gray!20}
 &  &  &  & AR: Request permission \tabularnewline
\rowcolor{gray!20}
 &  &  &  & XM2VTS: Pay a fee \tabularnewline
CUFSF \cite{zhang2011informationTheoretic,wang2009faceSketchRec} &  1,194 & Viewed & Hand drawn & Sketch: Free to download \tabularnewline
 &  &  &  & Photo: Request permission \tabularnewline
\rowcolor{gray!20}
IIIT-D viewed sketch \cite{bhatt2012memeticSketch} &  238 & Viewed & Hand drawn & Request permission \tabularnewline
IIIT-D semi-forensic sketch \cite{bhatt2012memeticSketch} &  140 & Semi-Forensic & Hand drawn & Request permission \tabularnewline
\rowcolor{gray!20}
IIIT-D forensic sketch \cite{bhatt2012memeticSketch} & 190 & Forensic & Hand drawn and Composite & Request permission \tabularnewline
\whline
\end{tabular}
}
}

\end{table*}

There are five commonly used datasets for benchmarking SBFR systems. Each contains pairs of sketches and photos. They differ by size, whether sketches are viewed and if drawn by artist or composited by software. Tab~\ref{Tab:existing-dataset} summaries each dataset in terms of these  attributes. 

CUHK Face sketch dataset (CUFS) \cite{wang2009faceSketchRec} is widely used in SBFR. It includes 188 subjects from the Chinese University of Hong Kong (CUHK) student dataset, 123 faces from the AR dataset \cite{Martinez1998}, and 295 faces from the XM2VTS dataset \cite{Messer99xm2vtsdb:the}. There are 606 faces in total. For each subject, a sketch and a photo are provided. The photo is taken of each subject with frontal pose and neutral expression under  normal lighting conditions. The sketch is then drawn by an artist based on the photo.


CUHK Face Sketch FERET Dataset (CUFSF) \cite{zhang2011informationTheoretic,wang2009faceSketchRec} is also commonly used to benchmark SBFR algorithms. There are 1,194 subjects from the FERET dataset \cite{879790}. For each subject, a sketch and a photo is also provided. However, compared to CUFS, instead of normal light condition, the photos in CUFSF are taken with lighting variation. Meanwhile, the sketches are drawn with shape exaggeration based on the corresponding photos. Hence, CUFSF is more challenging and closer to practical scenarios \cite{zhang2011informationTheoretic}.


The IIIT-D Sketch Dataset \cite{bhatt2012memeticSketch} is another well known facial sketch dataset. Unlike CUFS and CUFSF, it contains not only viewed sketches but also semi-forensic sketches and forensic sketches, therefore can be regarded as three separate datasets each containing a particular type of sketches, namely IIIT-D viewed, IIIT-D semi-forensic and IIIT-D forensic sketch dataset.
IIIT-D viewed sketch dataset comprises a total of 238 sketch-image pairs. The sketches are drawn by a professional sketch artist based on photos collected from various sources. It comprises of 67 sketch-image pairs from the FG-NET aging dataset\footnote{Downloadable at http://www-prima.inrialpes.fr/FGnet/html/home.html}, 99 sketch-digital image from Labeled Faces in Wild (LFW) dataset \cite{LFWTech}, and 72 sketch-digital image pairs from the IIIT-D student \& staff dataset \cite{LFWTech}. In the IIIT-D semi-forensic dataset, sketches are drawn based on an artist's memory instead of directly based on the photos or the description of an eye-witness. These sketches are termed semi-forensic sketches. The semi-forensic dataset is based on 140 digital images from the Viewed Sketch dataset. In the IIIT-D forensic dataset there are 190 forensic sketches and face photos. It contains 92 and 37 forensic sketch-photo pairs from \cite{Gibson2008} and  \cite{Taylor2001} respectively, as well as 61 pairs from various sources on the internet.
 
It is worth noting that the accessibility of these datasets varies, with some  not being publicly available. \cite{Klare2010} created a forensic dataset from sketches cropped from two books (also contained in IIIT-D forensic), which is thus limited by copyright. Klare et al. also conducted experiments querying against a real police database of 10,000 mugshots, but this is not publicly available.



\subsection{Viewed sketch face recognition}

Viewed sketch  recognition is the most studied sub-problem of SBFR. Although a hypothetical problem (in practice a photo would be used directly if available, rather than a viewed sketch), it provides an important step toward ultimately improving forensic sketch accuracy. It is hypothesized that based on an ideal eyewitness description, unviewed sketches would be equivalent to viewed ones.  Thus performance on viewed sketches should be an upper bound on expected performance on forensic sketches.


Viewed sketch-based face recognition studies can be classified into synthesis, projection and feature-based methods according to their main contribution to bridging the cross-modal gap.


\begin{figure}[t]
\centering
\subfigure[Match photo to sketch]{\includegraphics[height=1.8in]{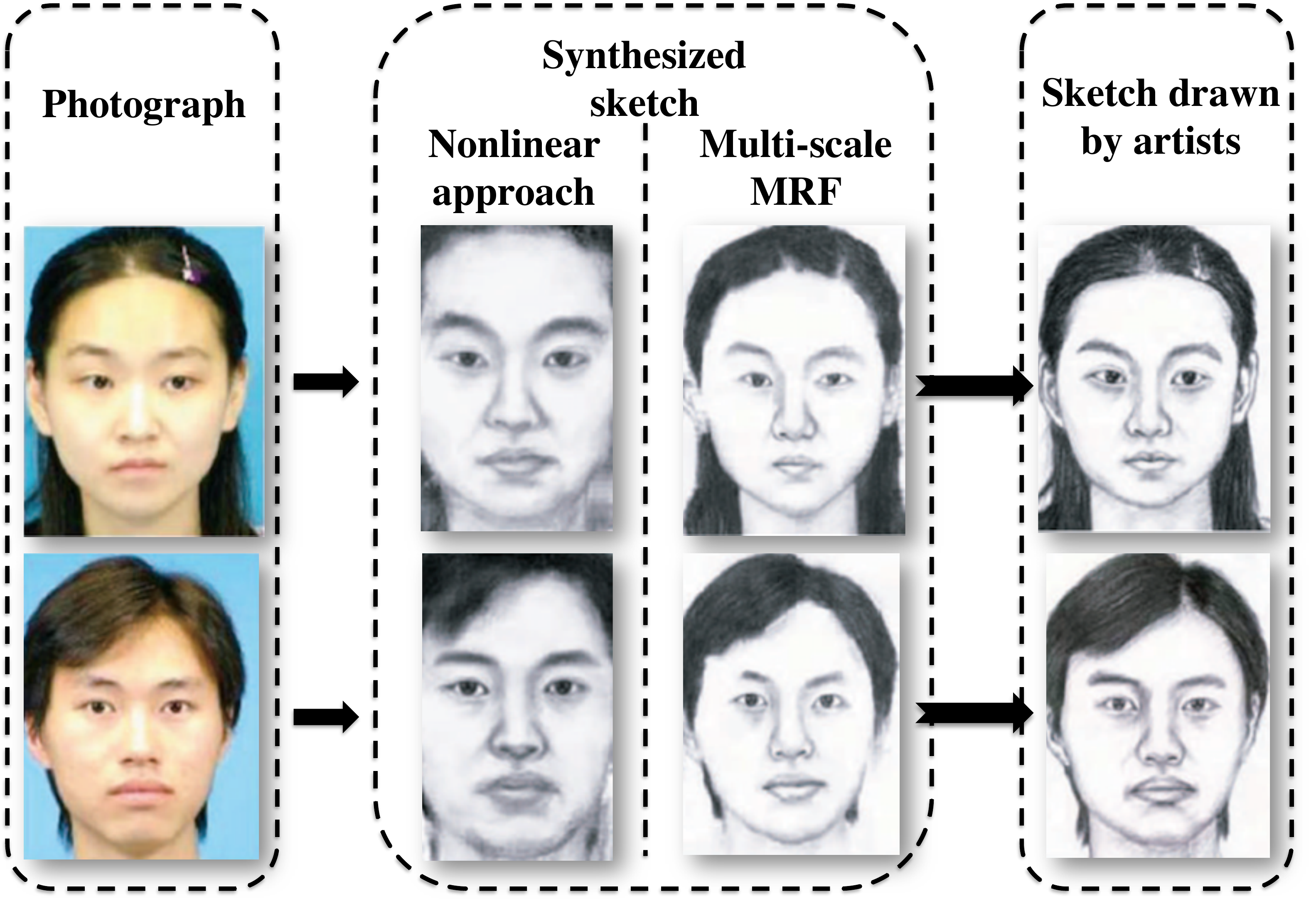}}
\hspace{0.1in}
\subfigure[Match sketch to photo]{\includegraphics[height=1.8in]{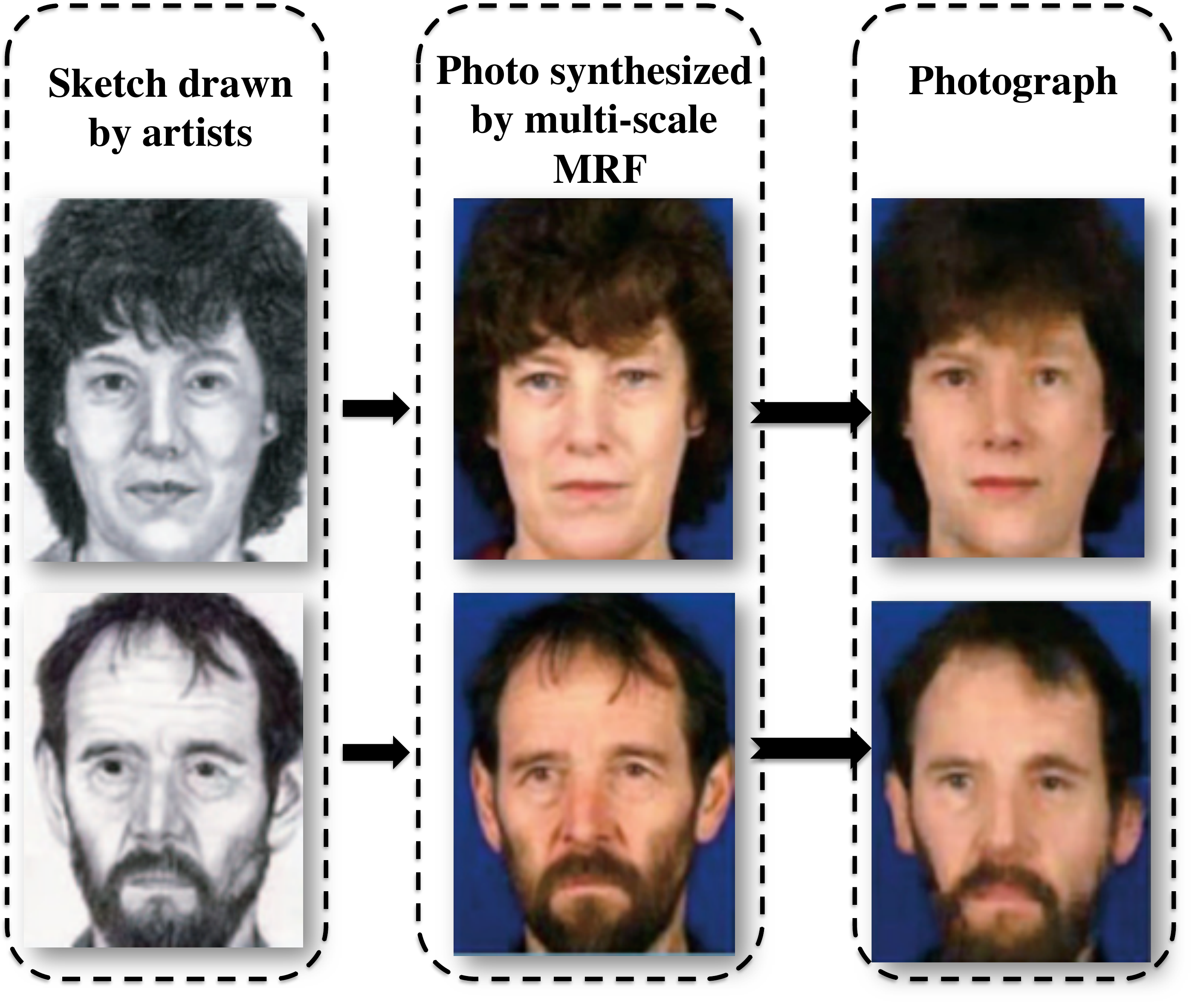}}
   \caption{Examples of sketch synthesis: (a) photo to sketch by synthesized sketches (b) sketch to photo by synthesized photos}
\label{fig:photo-to-sketch-synthsis}
\end{figure}


\subsubsection{Synthesis-based approaches}

The key strategy in synthesis-based approaches is to synthesize a photo from corresponding sketch (or vice-versa), after which traditional  homogeneous recognition methods can be applied (see Fig~\ref{fig:photo-to-sketch-synthsis}).  To convert a photo into a sketch, an eigensketch transformation algorithm is proposed by \cite{tang2002facePhotoSketch}. Classification is then accomplished by the obtained eigensketch features. To exploit the strong correlation exists among face images, the Karhunen-Loeve Transform (KLT) is applied to represent and recognise faces. The eigensketch transformation algorithm reduced the discrepancies between photo and sketch. The resulting rank-10 accuracy is reasonable. However, the work lacks in the small size of the dataset (188 pairs) used and weak  rank-1 accuracy.

Liu et al. \cite{1467376} proposed a  Local Linear Embedding (LLE) inspired method to convert photos into sketches based on image patches. Those sketches are geometry preserving synthetic sketches. For each image patch to be converted, it finds the nearest neighbors in the training set, and uses their corresponding sketch patches to synthesize the sketch patch.
Tang and Wang \cite{wang2009faceSketchRec} further improved \cite{1467376} by developing an approach to synthesize local face structures at different scales using a Markov Random Fields (MRF), as shown in Fig~\ref{fig:photo-to-sketch-synthsis}(a). In the latter work, a multi-scale MRF learns local patches and scales jointly instead of independently as in \cite{1467376}. The scale of learned local face structures is based on the size of overlapped patches. With a multi-scale MRF, the joint photo-sketch model is learned at multiple scales. By converting face photos or sketches into same modality, the modality-gap is reduced, thus allowing the two domains to be matched effectively.

In both \cite{tang2002facePhotoSketch} and \cite{wang2009faceSketchRec}, after photos/sketches are synthesized, many standard methods like PCA~\cite{turk1991eigenfaces}, Bayesianface~\cite{598227}, Fisherface~\cite{598228}, null-space LDA~\cite{Chen20001713}, dual-space LDA~\cite{1315214} and Random Sampling LDA (RS-LDA)~\cite{1315172,Wang06randomsampling} are straightforwardly applied for homogeneous face recognition.

The embedded hidden Markov model (E-HMM) is applied by Zhong et al.~\cite{4217122}  to transform a photo to a sketch. The nonlinear relationship between a photo/sketch pair is modeled by E-HMM. Then, learned models are used to generate a set of pseudo-sketches. Those pseudo-sketches are used to synthesize a finer face pseudo-sketch based on a selective ensemble strategy. E-HMMs are also used by Gao et al.~\cite{Gao2008,4453838} to synthesis sketches from photos. On the contrary, Xiao et al.~\cite{Xiao20091576} proposed a E-HMM based method to synthesis photos from sketches. Liu et al.~\cite{Liu:2007:BTI:1625275.1625621} proposed a synthesis method based on Bayesian Tensor Inference. This method can be used to synthesize both sketches from photos and photos from sketches.

\subsubsection{Projection based approaches}

Rather than trying to completely reconstruct one modality from the other as in synthesis-based approaches; projection-based approaches attempt to find a lower-dimensional sub-space in which the two modalities are directly comparable (and ideally, in which identities are highly differentiated).

Lin and Tang~\cite{lin2006interModalityFace} proposed a linear transformation which can be used between different modalities (sketch/photo, NIR/VIS), called common discriminant feature extraction (CDFE). In this method, images from two modalities are projected into a common feature space in which matching can be  effectively performed.


Sharma et al.~\cite{sharma2011pls} use Partial Least Squares (PLS) to linearly map images of different modalities (e.g., sketch, photo and different poses, resolutions) to a common subspace where mutual covariance is maximized. This is shown to generalize better than CCA. Within this subspace, final matching is performed with simple NN.

In \cite{huang2013coupledDictionary}, a unified sparse coding-based model for coupled dictionary and feature space learning is proposed to simultaneously achieve synthesis and recognition in a common subspace. The learned common feature space is used to perform cross-modal face recognition with NN.

In \cite{1467376} a kernel-based nonlinear discriminant analysis (KNDA) classifier is adopted by Liu et al. for sketch-photo recognition. The central contribution is to use the nonlinear kernel trick to map input data into an implicit feature space. Subsequently, LDA is used to extract features in that space, which are non-linear discriminative features of the input data.

\subsubsection{Feature based approaches}

Rather mapping photos into sketches, or both into a common subspace; feature-based approaches focus on designing a feature descriptor for each image that is intrinsically invariant to the modality, while being variant to the identity of the person.  The most widely used image feature descriptors are Scale-invariant feature transform (SIFT), Gabor transform, Histogram of Averaged Oriented Gradients (HAOG) and Local Binary Pattern (LBP). Once sketch and photo images are encoded using these descriptors, they may be matched directly, or after a subsequent projection-based step as in the previous section.

Klare et al.~\cite{Klare2010} proposed the first direct sketch/photo matching method based on invariant SIFT-features \cite{Lowe:2004aa}. SIFT features provide a compact vector representation of an image patch based on the magnitude, orientation, and spatial distribution of the image gradients \cite{Klare2010}. SIFT feature vectors are first sampled uniformly from the face images and concatenated together separately for sketch and photo images. Then, Euclidean distances are computed between concatenated SIFT feature vectors of sketch and photo images for NN matching.

Later on, Bhatt et al.~\cite{5634507} proposed an method which used extended uniform circular local binary pattern  descriptors to tackle  sketch/photo matching. Those descriptors are based on discriminating facial patterns formed by high frequency information in facial images. To obtained  the high frequency cues, sketches and photos are decomposed into multi-resolution pyramids. After extended uniform circular local binary pattern based descriptors are computed, a Genetic Algorithm (GA) \cite{Goldberg:1989:GAS:534133} based weight optimization technique is used to find optimum weights for each facial patch. Finally, NN matching is performed by using weighted Chi square distance measure.

Khan et al.~\cite{6411670} proposed a self-similarity descriptor. Features are extracted independently from local regions of sketches and photos. Self-similarity features are then obtained by correlating a small image patch within its larger neighborhood. Self-similarity remains relatively invariant to the photo/sketch-modality variation therefore reduces the modality gap before NN matching.

A new face descriptor, Local Radon Binary Pattern (LRBP) was proposed by Galoogahi et al.~\cite{6467240} to directly match face photos and sketches. In the LRBP framework, face images are first transformed into Radon space, then transformed face images are encoded by Local Binary Pattern (LBP). Finally, LRBP is computed by concatenating histograms of local LBPs. Matching is performed by a distance measurement based on Pyramid Match Kernel (PMK) \cite{1641019}. LRBP benefits from low  computational complexity and the fact that there is no critical parameter to be tuned \cite{6467240}.

Galoogahi et al. consequently proposed another two face descriptors: Gabor Shape \cite{KianiGaloogahi:2012:FPR:2393347.2396354} which is variant of Gabor features and Histogram of Averaged Oriented Gradient (HAOG) features \cite{galoogahi2012interModalityFace} which is variant of HOG for sketch/photo directly matching, the latter achieves perfect $100\%$ accuracy on the CUFS dataset.

\begin{table*}[t]
\tbl{Sketch-Photo matching methods: Performance on benchmark datasets.\label{Tab:feature-based-methods-d1}}{%
\renewcommand{\arraystretch}{1.5}
\scalebox{0.65}{
\begin{tabular}{lclllcc}
\whline
Method & Publications & Recognition Approach & Dataset & Feature & Train:Test & Accuracy \tabularnewline
\hline
\rowcolor{gray!20}
Synthesis based & \cite{tang2002facePhotoSketch} & KLT  & CUHK & Eigen-sketch features & 88:100  &  about 60\% \tabularnewline
\rowcolor{gray!20}
  & \cite{1467376} & KNDA  & CUFS &  & 306:300  &  87.67\% \tabularnewline
\rowcolor{gray!20}
  &  \cite{wang2009faceSketchRec} & RS\_LDA & CUFS & Multiscale MRF & 306:300  &  96.3\% \tabularnewline
\rowcolor{gray!20}
  &  \cite{4217122} &  & CUFS & E-HMM &  ---  &  95.24\% \tabularnewline
\rowcolor{gray!20}
  &  \cite{Liu:2007:BTI:1625275.1625621} &  & CUFS & E-HMM$+$Selective ensemble &  ---  &  100\% \tabularnewline
Projection based & \cite{Klare2010} & Common representation & CUFS & SIFT & 100:300 &96.47\%  \tabularnewline
 &  \cite{sharma2011pls} & PLS & CUHK & & 88:100 & 93.60\%  \tabularnewline
 & \cite{choi2012cvpr} & PLS regression & CUFS,CUFSF & Gabor and CCS-POP & 0:1800 & 99.94\% \tabularnewline 
\rowcolor{gray!20}
Feature based & \cite{Klare2010} & NN & CUFS & SIFT & 100:300  & 97.87\% \tabularnewline
\rowcolor{gray!20}
 & \cite{6411670} & NN & CUFS & Self Similarity  &161:150 & 99.53\%  \tabularnewline
\rowcolor{gray!20}
 & \cite{6467240} & NN,PMK,Chi-square  & CUFS  & LRBP & --- & 99.51\%  \tabularnewline
\rowcolor{gray!20}
 & \cite{KianiGaloogahi:2012:FPR:2393347.2396354} & NN,Chi-square & CUFS & Gabor Shape & 306:300 & 99.14\% \tabularnewline
  \rowcolor{gray!20}
 & \cite{5634507} & Weighted Chi-square & CUFS & EUCLBP & 78:233 & 94.12\%  \tabularnewline
\rowcolor{gray!20}
 & \cite{galoogahi2012interModalityFace} & NN,Chi-square & CUFS & HAOG & 306:300 & 100.00\%  \tabularnewline
  \rowcolor{gray!20}
 & \cite{KianiGaloogahi:2012:FPR:2393347.2396354} & NN,Chi-square & CUFSF & Gabor Shape & 500:694 & 96.32\% \tabularnewline
  \rowcolor{gray!20}
 & \cite{6467240} & NN,PMK,Chi-square  & CUFSF  & LRBP & --- & 91.12\%  \tabularnewline
\rowcolor{gray!20}
 & \cite{pramanik2012geo} & K-NN & CUHK & Geometric features & 108:80 & 80.00\% \tabularnewline
\rowcolor{gray!20}
 & \cite{5634507} & Weighted Chi-square & IIIT-D & EUCLBP & 58:173 & 78.58\% \tabularnewline
\whline
\end{tabular}
}
}

\end{table*}

Klare et al.~\cite{Klare2010} further exploited their SIFT descriptor, by combining it with a `common representation space' projection-based strategy. The assumption is that even if sketches and photos are not directly comparable, the distribution of \emph{inter-face similarities} will be similar within the sketch and photo domain. That is, the (dis)similarity between a pair of sketches will be roughly the same as the (dis)similarity between the corresponding pair of photos. Thus each sketch and photo is re-encoded as a vector of their euclidean distances to the training set of sketches and photos respectively. This common representation should now be invariant to modality and sketches/photos can be compared directly. To further improve the results, direct matching and common representation matching scores are fused to generate the final match \cite{Klare2010}. The advantage of this approach over mappings like CCA and PLS is that it does not require the sketch-photo domain mapping to be linear. The common representation strategy has also been used to achieve cross-view person recognition \cite{an2013referenceReID}, where it was shown to be dependent on sufficient training data.

In contrast to the previous methods which are appearance centric in their representation, Pramanik et al.~\cite{pramanik2012geo} evaluate an analytic geometry feature based recognition system. Here, a set of facial components such as eyes, nose, eyebrows, lips, are extracted their aspect ratio are encoded as feature vectors, followed by K-NN as classifier.


\subsection{Forensic sketch face recognition}

Forensic sketches pose greater challenges than viewed sketch recognition because forensic sketches contain less, incomplete or inaccurate information. This issue due to the subjectivity of the description, and imperfection of the witness' memory.

There are therefore two sets of challenges in forensic sketch-based  recognition: (1) recognizing across modalities and (2) performing recognition despite inaccurate, incomplete and harder to align depictions of the face. Due to its greater challenge, and the lesser availability of forensic sketch datasets, research in this area has been less than for viewed sketches. Uhl et al.~\cite{517132} proposed the first system for automatically matching police artist sketches to photographs. In their method, facial features are first extracted from sketches and photos. Then, the sketch and photo are geometrically standardized to facilitate comparison. Finally, eigen-analysis is employed for matching. Only 7 probe sketches were used in experimental validation, their method is antiquated with respect to modern methods. Nonetheless, Uhl and Lobo's study highlighted the complexity and difficulty in forensic sketch based face recognition and drew other researchers towards forensic sketch-based face recognition.

Klare et al.~\cite{5601735} performed the first large scale study in 2011, with an approach combining feature-based and projection-based contributions. SIFT and MLBP features were extracted, followed by training a LFDA projection to minimize the distance between corresponding sketches and photos while maximizing the distance between distinct identities. They analyse a dataset of 159 pairs of forensic hand drawn sketches and mugshot photos. The subjects in this dataset were identified by the law enforcement agencies. They also included 10,159 mugshot images provided by Michigan State Police to better simulate a realistic police search against a large gallery. With this realistic scenario, they achieved about 15 percent success rate. 

To improve recognition performance, Bhatt et al.~\cite{bhatt2012memeticSketch} proposed an algorithm that also combines feature and projection-based contributions. They use multi-scale circular Webber's Local descriptor to encode structural information in local facial regions. Memetic optimization was then applied to every local facial region as a metric learner to find the optimal weights for Chi squared NN matching \cite{bhatt2012memeticSketch}. The result outperforms \cite{5601735} using only the forensic set as gallery. 

\subsection{Composite sketches based face recognition}

Three studies have thus far focused on face recognition using composite sketches. The first one uses both local and global features to represent sketches and is proposed by Yuen et al.~\cite{4244565}. This method required user input in the form of relevance feedback in the recognition phase. The second two focus on holistic \cite{6330967} and component based \cite{Han2013} representations respectively.

The holistic method \cite{6330967} uses similarities between local features computed on uniform patches across the entire face image. Following tessellating a facial sketch/mugshot into 154 uniform patches, SIFT \cite{Lowe:2004aa} and multi-scale local binary pattern (MLBP) \cite{1017623} invariant features are extracted from each patch. With this feature encoding, as improved version of the  common representation intuition from \cite{Klare2010} is  applied, followed by RS-LDA \cite{raey} to generate a discriminative subspace for NN matching with cosine distance. The scores generated by each feature and patch are fused for final recognition.


In contrast, the component based method \cite{Han2013} uses similarities between individual facial components to compute an overall sketch to mughsot match score.
Facial landmarks in composite sketches and  photos are automatically detected  by an active shape model (ASM) \cite{Milborrow:2008:LFF:1478237.1478275}. Mutiscale local binary patterns (MLBPs) are then applied to extract features of each facial component, and similarity is calculated for each component: using histogram intersection distance for the component's appearance and cosine distance for its shape. The similarity scores of each facial component are normalized and fused to obtain the overall sketch-photo similarity. 


\subsection{Caricature based face recognition}

\begin{figure}[t]
  \centering
  \includegraphics[height = 2.3in]{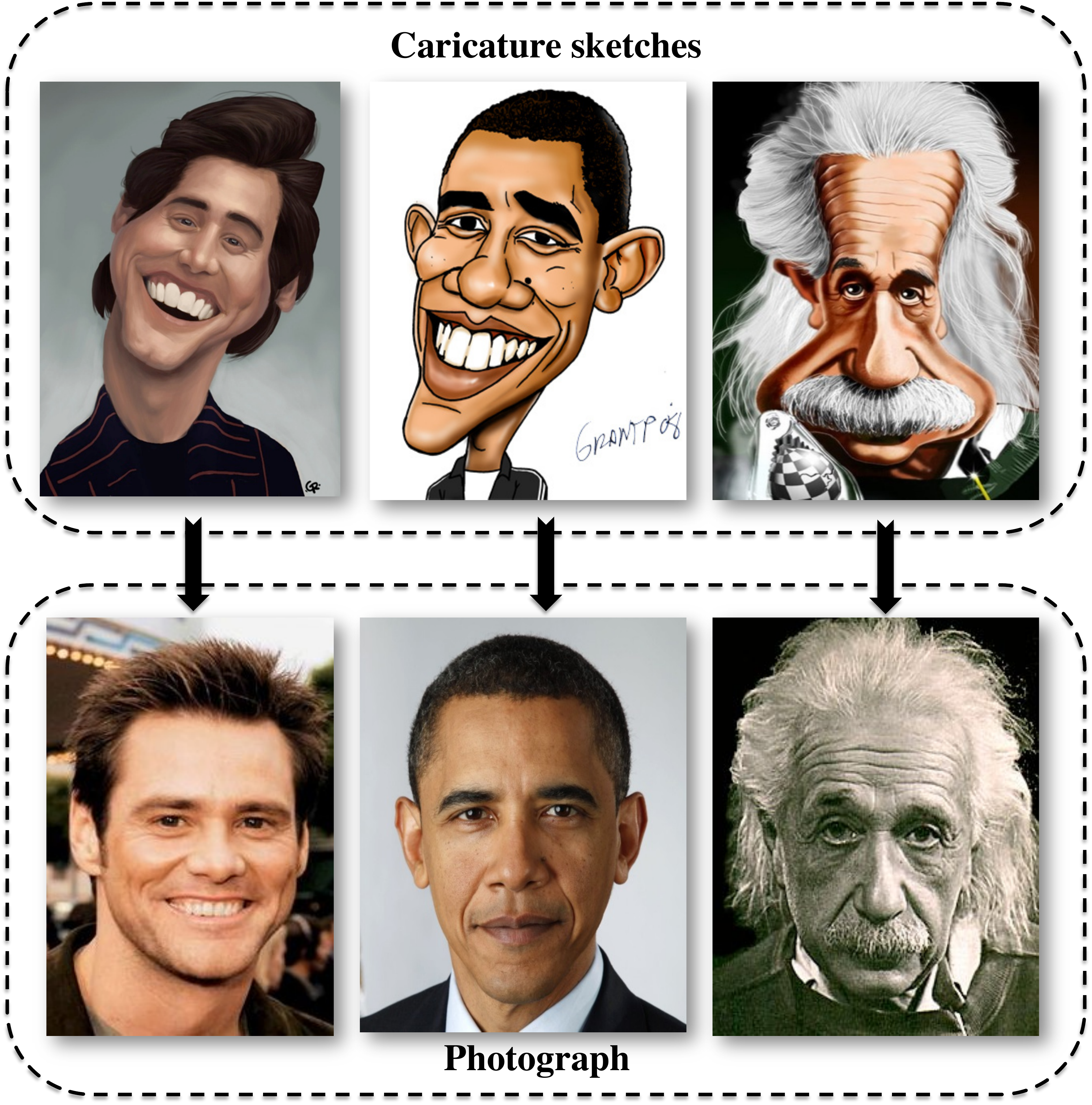}
  \caption{Caricatures and corresponding mugshots}
  \label{fig:caricature-mug}
\end{figure}

The human visual system's ability to recognise a person from a caricature is remarkable, as conventional face recognition approaches fail in this setting of extreme intra-class variability (Fig~\ref{fig:caricature-mug}). The caricature generation process can be conceptualised as follows: If we assume a face space in which each face lies. Then by drawing a line to connect the mean face to each face, the corresponding caricature will lie beyond that face along the line. That is to say, a caricature is an exaggeration of a face away from the mean \cite{Lanckriet:2004:LKM:1005332.1005334}. 

Studies have suggested that people may encode faces in a caricatured manner \cite{sinha2006humanRecognition}. Moreover they may be more capable of recognizing a familiar person through a caricature than an accurate rendition \cite{Mauro1992,Rhodes1987}. The effectiveness of a caricature is due to its emphasis of deviations from average faces \cite{klare2012charicature}. Developing efficient approaches in caricature based face recognition could help drive more robust and reliable face and heterogeneous face recognition systems.

Klare et al.~\cite{klare2012charicature} proposed a semi-automatic system to match caricatures to photographs. In this system, they defined a set of qualitative facial attributes that describe the appearance of a face independently of whether it is a caricature or photograph. These mid-level facial features were manually annotated for each image, and used together with automatically extracted LBP \cite{1017623} features. These two feature types were combined with an ensemble of matching methods including NN and discriminatively trained logistic regression SVM, MKL and LDA.  The results showed that caricatures can be recognized slightly better with high-level qualitative features than low-level LBP features, and that they are synergistic in that combining the two can almost double the performance up to $22.7\%$ rank 1 accuracy. A key insight here is that -- in strong contrast to viewed sketches that are perfectly aligned -- the performance of holistic feature based approaches is limited because the exaggerated nature of caricature sketches means that detailed alignment is impossible.

A limitation of the above work is that the facial attributes must be provided, requiring manual intervention at run-time. Ouyang et al.~\cite{sxoyBeyong} provided a fully automated procedure that uses a classifier ensemble to robustly estimate facial attributes separately in the photo and caricature domain. These estimated facial attributes are then combined with low-level features using CCA to generate a robust domain invariant representation that can be matched directly.  This study also contributed facial attribute annotation datasets that can be used to support this line of research going forward.

\subsection{Summary and Conclusions}

Tab~\ref{Tab:feature-based-methods-d1}  summarizes the results of major studies in terms of distance metric, dataset, feature representation, train to test ratio, and rank-1 accuracy, of feature-based and projection-based approaches respectively. As viewed sketch datasets exhibit near perfect alignment and detail correspondence between sketches and photos, well designed approaches achieve near perfect accuracies. Note that some results on the same dataset are not directly comparable because of differing test set sizes.

\paragraph{Methodologies}
All three categories of approaches -- synthesis, projection and discriminative features -- have been well studied for SBFR. Interestingly, while synthesis approaches have been one of the more popular categories of methods, they have only been demonstrated to work in viewed-sketch situations where the sketch-photo transformation is very simple and alignment is perfect. It seems unlikely that they can generalize effectively to forensic sketches, where the uncertainty introduced by forensic process (eyewitness subjective memory) significantly completes the matching process.

An interesting related issue that has not been systematically explored by the field is the dependence on the sketching artists. Al Nizami et al.~\cite{5339043} demonstrated significant intra-personal variation in sketches drawn by different artists. This may challenge systems that rely on learning a simple uni-modal cross- modal mapping. This issue will become more significant in the forensic sketch case where there is more artist discretion, than in viewed-sketches which are more like copying exercises. 

\paragraph{Challenges and Datasets}
The majority of SBFR research has focused on viewed sketch-based recognition, with multiple studies now achieving near-perfect results on the CUFS dataset. This is due to the fact that viewed sketches are professionally rendered copies of photographed faces, and thus close in likeness to real faces, so non-linear misalignment and all the attendant noise introduced by verbal descriptions communicated from memory are eliminated. This point is strongly made by Choi et al.~\cite{choi2012cvpr}, who criticize the existing viewed-sketch datasets and the field's focus on them. They demonstrate that with minor tweaks, an off the shelf PLS-based \emph{homogeneous} face recognition system can outperform existing cross-modality approaches and achieve perfect results on the CUFS dataset. They conclude that existing viewed-sketch datasets are unrealistically easy, and not representative of realistic forensic sketch scenarios. 

It is thus important that the field should move to more challenging forensic, composite and caricature sketches with more realistic non-linear misalignment and heteroskedastic noise due to the forensic process. This will reveal  whether current state of the art methods from viewed-sketches are indeed best, or are brittle to more realistic data; and will drive the generation of new insights, methods and practically relevant capabilities. Research here, although less mature, has begun to show promising results. However, it is being hampered by lack of readily obtainable forensic datasets. Constructing realistic and freely available datasets should be a priority \cite{choi2012cvpr}. 

\paragraph{Training Data Source} Many effective SBFR studies have leveraged annotated training data to learn projections and/or classifiers \cite{Klare2010}.  As interest has shifted onto forensic sketches, standard practice has been to train such models on viewed-sketch datasets and test on forensic datasets \cite{5601735}. An interesting question going forward is whether this is the right strategy. Since viewed-sketches under-represent sketch-photo heterogeneity, this means that learning methods are learning a model that is not matched to the data (forensic sketches) that they will be tested on. This poses an additional challenge of \emph{domain shift} \cite{pan2009transfer_survey}, to be solved; as well as further motivating the creation of larger forensic-sketch datasets with which it will be possible to discover whether training on forensic pairs is more effective than training on viewed-pairs.

\section{Matching NIR to Visible Light Images}\label{nir}

\begin{figure}[t]
\begin{center}
\includegraphics[height = 2.3in]{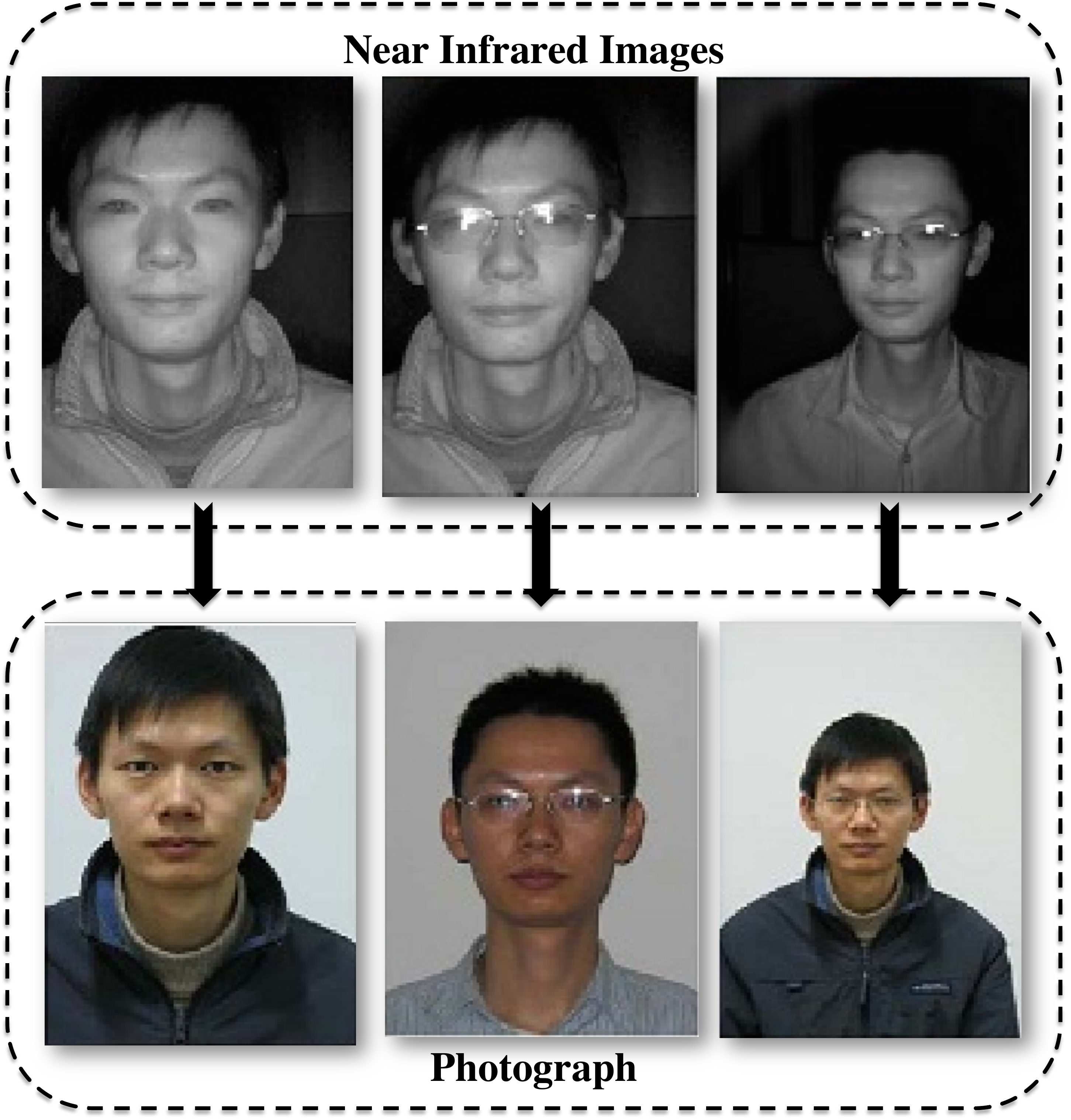}
\end{center}
   \caption{VIS and NIR face images.}
\label{fig:nir}
\end{figure}

NIR face recognition has attracted increasing attention recently because of its much desired attribute of (visible-light) illumination invariance, and the decreasing cost of NIR acquisition devices. It encompasses matching near infrared (NIR) to visible light (VIS) face images. In this case, the VIS enrollment samples are images taken under visible light spectrum (wavelength range $0.4\mu m-0.7\mu m$), while query images are captured under near infrared (NIR) condition (just beyond the visible light range, wavelengths between $0.7\mu m$ -  $1.4\mu m$) \cite{5597000}. NIR images are close enough to the visible light spectrum to capture the structure of the face, while simultaneously being far enough to be invariant to visible light illumination changes. Fig~\ref{fig:nir} illustrates differences between NIR and VIS images. Matching NIR to VIS face images is of interest, because it offers the potential for face recognition where controlling the visible environment light is difficult or impossible, such as in night-time surveillance or automated gate control.

In NIR based face recognition, similar to sketch based recognition, most studies can be categorized into synthesis, projection and discriminant feature based approaches, according to their contribution to bridging the cross-modal gap. 

\subsection{Datasets}

There are four main heterogeneous datasets covering the NIR-VIS condition. 
The CASIA HFB dataset \cite{li2009hfb}, composed of visual (VIS), near infrared (NIR) and 3D faces, is widely used. In total, it includes 100 subjects: 57 males and 43 females. For each subject, there are 4 VIS and 4 NIR face images. Meanwhile, there are also 3D images for each subject (92 subjects: 2 for each, 8 subjects: 1 for each). In total, there are 800 images for NIR-VIS setting and 200 images for 3D studies.

CASIA NIR-VIS 2.0 \cite{li2013nirDB2} is another widely used NIR dataset. 725 subjects are included, with 50 images (22 VIS and 28 NIR) per subject, for a total of 36,250 images.

The Cross Spectral Dataset \cite{goswami2011} is proposed by Goswami et al. It consists of 430 subjects from various ethnic backgrounds (more than 20\% of non-European origin).  At least one set of 3 poses (-10 degree / 0 degree / 10 degree) are captured for each subject. In total, there are 2,103 NIR images and 2,086 VIS images.

The PolyU NIR face dataset \cite{Zhang20102337} is proposed by the biometric research center at Hong Kong Polytechnic University. This dataset includes 33,500 images from 335 subjects. Besides frontal face images and faces with expression, pose variations are also included. The active light source used to create this dataset is in the NIR spectrum between 780mm to 1,100mm.

A summary of the main NIR-VIS datasets can be found in Tab~\ref{Tab:existing-dataset2}. Each column categorizes the datasets by wavelength of NIR light, no. of subject, no. of images, and whether  they include 3D images, pose and expression variations, respectively.

\begin{table*}
\tbl{Summary of existing NIR-VIS benchmark datasets\label{Tab:existing-dataset2}}{%
\renewcommand{\arraystretch}{1.5}
\scalebox{0.7}{
\begin{tabular}{lcccccccc}
\whline
Dataset & Wavelength & No.of Subjects & No.of Images &  3D & Pose variations & Expression variations\tabularnewline
\hline
\rowcolor{gray!20}
CASIA HFB \cite{li2009hfb} & 850nm & 100 & 992 & $\surd$ &  $\times$ & $\times$ \tabularnewline
CASIA NIR-VIS 2.0 \cite{li2013nirDB2} & 850nm & 725 & 17580 & $\surd$ &  $\surd$ & $\surd$ \tabularnewline
\rowcolor{gray!20}
Cross Spectral Dataset \cite{Zhang20102337} & 800-1000nm & 430 & 4189 & $\surd$ &  $\surd$ & $\times$ \tabularnewline
PolyU \cite{Zhang20102337} & 780-1100nm & 335 & 33500 & $\surd$ &  $\surd$ & $\surd$ \tabularnewline
\whline
\end{tabular}
}
}
\label{Tab:existing-dataset2}
\end{table*}

\subsection{Synthesis based approaches}

Wang et al.~\cite{Wang2009} proposed an analysis-by-synthesis framework, that transforms face images from NIR to VIS. To achieve the conversion, facial textures are extracted from both modalities. NIR-VIS texture patterns extracted at corresponding regions of different face pairs collectively compose a training set of matched pairs. After illumination normalization \cite{Xie2006609}, VIS images can be synthesized patch-by-patch by finding the best matching patch for each patch of the input NIR image.

Chen et al.~\cite{chen2009cvpr} also synthesize VIS  from NIR images using a similar inspiration of learning a cross-domain dictionary of corresponding VIS and NIR patch pairs. To more reliably match patches, illumination invariant LBP features are used to represent them. Synthesis of the VIS image is further improved compared to \cite{Wang2009}, by using locally-linear embedding (LLE) inspired patch synthesis rather than simple nearest-neighbor. Finally homogeneous VIS matching is performed with NN classifier on the LBP representations of the synthesized images.

Xiong et al.~\cite{pengfei2012} developed a probabilistic statistical model of the mapping between two modalities of facial appearance, introducing a hidden variable to represent the transform to be inferred. To eliminate the influences of facial structure variations, a 3D model is used to perform pose rectification and pixel-wise alignment. Difference of Gaussian (DOG) filter is further used to normalize image intensities.

\subsection{Projection based approaches}

Lin et al.~\cite{lin2006interModalityFace}  proposed a matching method based on Common Discriminant Feature Extraction (CDFE), where two linear mappings are learned to project the samples from NIR  and VIS modalities to a common feature space. The optimization criterion aims to both minimize the intra-class scatter while maximizing the inter-class scatter. 
They further extended the algorithm to deal with more challenging situations where the sample distribution is non-gaussian by kernelization, and where the transform is multi-modal.

After analysing the properties of NIR and VIS images, Yi et al.~\cite{yi2007nirVisFaceCCA} proposed a learning-based approach for cross-modality matching. In this approach, linear discriminant analysis (LDA) is used to extract features and reduce the dimension of the feature vectors. Then, a canonical correlation analysis (CCA) \cite{Hotelling:1992aa} based mechanism is learned to project feature vectors from both modalities into CCA subspaces. Finally, nearest-neighbor with cosine distance is used matching score.

Both of methods proposed by Lin and Yi tend to overfit to training data. To overcome this limitation, Liao et al.~\cite{liao2009hfr} present a algorithm based on learned intrinsic local image structures. In training phase, Difference-of-Gaussian filtering is used to normalize the appearance of heterogeneous face images in the training set. Then, Multi-scale Block LBP (MB-LBP) \cite{Liao:2007aa} is applied to represent features called Local Structure of Normalized Appearance (LSNA). The resting representation is high-dimensional, so Adaboost is used for feature selection to discover a subset of informative features. R-LDA is then applied on the whole training set to construct a discriminative subspace. Finally, matching is performed with a verification-based strategy, where cosine distance between the projected vectors is compared with a threshold to decide a match.

Klare et al.~\cite{5597000} build on \cite{liao2009hfr}, but improve it in a few ways. They add HOG to the previous LBP descriptors to better represent patches, and use an ensemble of random LDA subspaces \cite{5597000} learn a shared projection with reduced over fitting. Finally, NN and Sparse Representation based matching are performed for matching.

Lei et al.~\cite{5206860} presented a method to match NIR and VIS face images called Coupled Spectral Regression(CSR). Similar to other projection-based methods, they use two  mappings to project the heterogeneous data into a common feature subspace. In order to further improve the performance of the algorithm (efficiency and generalisation), they use the solutions  derived from the view of graph embedding \cite{4016549} and spectral regression \cite{4408855} combined with regularization techniques. They later improve the same framework \cite{lei2012improvedCSR}, to better exploit the cross-modality supervision and sample locality.

Huang et al.~\cite{huang2013dsr} proposed a discriminative spectral regression (DSR) method that maps NIR/VIS face images into a common discriminative subspace in which robust classification can be achieved. They transform the subspace learning problem into a least squares problem. It is asked that images from the same subject should be mapped close to each other, while these from different subjects should be as separated as possible. To reflect category relationships in the data, they also developed two novel regularization terms.

\subsection{Feature based approaches}


Zhu et al.~\cite{zhu2013lgh} interpret the VIS-NIR problem as a highly illumination-variant task. They address it by designing an effective illumination invariant descriptor, the logarithm gradient histogram (LGH). This outperforms the LBP and SIFT descriptors used by \cite{liao2009hfr} and \cite{5597000} respectively. As a purely feature-based approach, no training data is required.

Huang et al.~\cite{huang2013mif}, in contrast to most approaches, perform feature extraction after CCA projection. CCA is used to maximize the correlations between NIR and VIS image pairs. Based on low-dimensional representations obtained by CCA, they extract three different modality-invariant features, namely, quantized distance vector (QDV), sparse coefficients (SC), and least square coefficients (LSC). These features are then represented with a sparse coding framework, and sparse coding coefficients are used as the encoding for matching.

Goswami et al.~\cite{goswami2011} introduced a new dataset for NIR/VIS (VIS/NIR) face recognition. To establish baselines for the new dataset they compared a series of photometric normalization techniques, followed by LBP-based encoding and LDA to find an invariant subspace. They compared classification with Chi-squared and Cosine as well as establishing a logistic-regression based verification model that obtained the best performance by fusing the weights from each of the model variants.

Gong and Zheng~\cite{gong2013acpr} proposed a learned feature descriptor, that adapts parameters to maximize the correlation of the encoded face images between two modalities. With this descriptor, the within-class variations can be reduced at the feature extraction stage, therefore offering better recognition performance. This descriptor outperforms classic HOG, LBP and MLBP, however unlike the others it requires training.

Finally, Zhu et al.~\cite{zhu2014transduction} presented a new logarithmic Difference of Gaussians (Log-DoG) feature, derived based on mathematical rather than merely empirical analysis of various features properties for recognition. Beyond this, they also present a framework for projecting to a non-linear discriminative subspace for recognition. In addition to aligning the modalities, and regularization with a manifold, their projection strategy uniquely exploits the unlabelled  test data transductively.


\subsection{Summary and Conclusions}

Given their decreasing cost, NIR acquisition devices are gradually becoming an integrated component of everyday surveillance cameras. Combined with the potential to match people in a (visible-light) illumination independent way, this has generated increasing interest in NIR-VIS face recognition. 

As with all the HFR scenarios reviewed here, NIR-VIS studies have addressed bridging the cross-modal gap with a variety of synthesis, projection and feature-based techniques. One notable unique aspect of NIR-VIS is that it is the change in illumination type that is the root of the cross-modal challenge. For this reason image-processing or physics based photometric normalization methods (e.g., gamma correction, contrast equalization, DoG filtering) often play a greater role in bridging the gap. This is because it is to some extent possible to model the cross-modal lighting gap more directly and explicitly than other HFR scenarios that rely entirely on machine learning or invariant feature extraction methods.

\section{Matching 2D to 3D}\label{sec:2d3d}

The majority of prior HFR systems work with 2D images, whether the face is photographed, sketched or composited. Owning to the 2D projection nature of these faces, such systems often exhibit high sensitivity to illumination and pose. Thus 3D-3D face matching has been of interest for some time \cite{bowyer2006faceRec2d3d}. However, 3D-3D matching is hampered in practice by the complication and cost of 3D compared to 2D equipment. An interesting variant of interest is thus the cross-modal middle ground, of using 3D images for enrollment, and 2D images for probes. This is useful, for example, in access control where enrollment is centralized (and 3D images are easy to obtain), but the access gate can be deployed with simpler and cheaper 2D equipment. In this case, 2D probe images can potentially be matched more reliably against the 3D enrollment model than a 2D enrollment image -- if the cross-domain matching problem can be solved effectively.


\subsection{Datasets}

\begin{figure}[t]
\begin{center}
\includegraphics[width=0.7in]{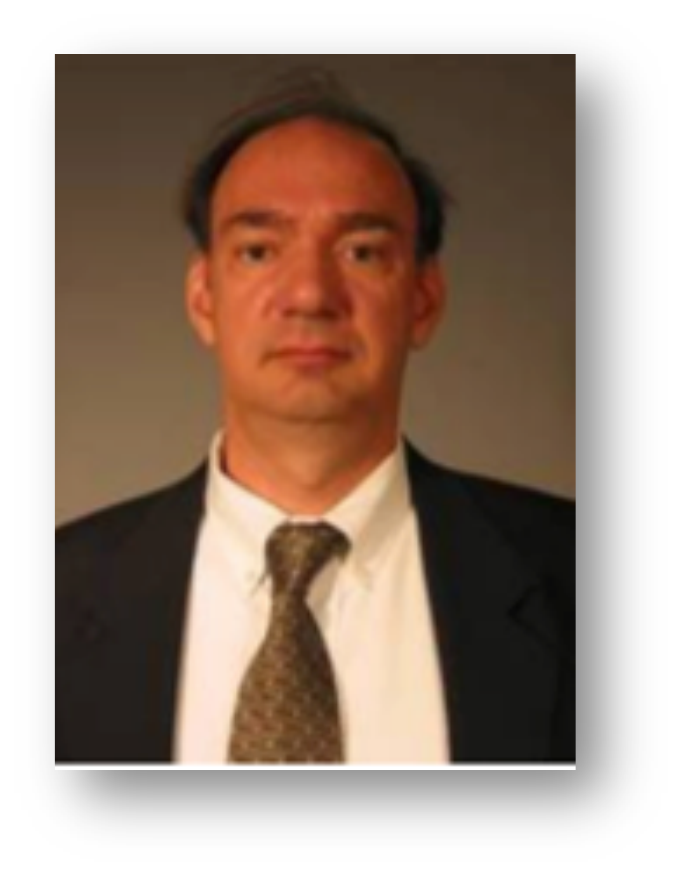}%
\includegraphics[width=0.7in]{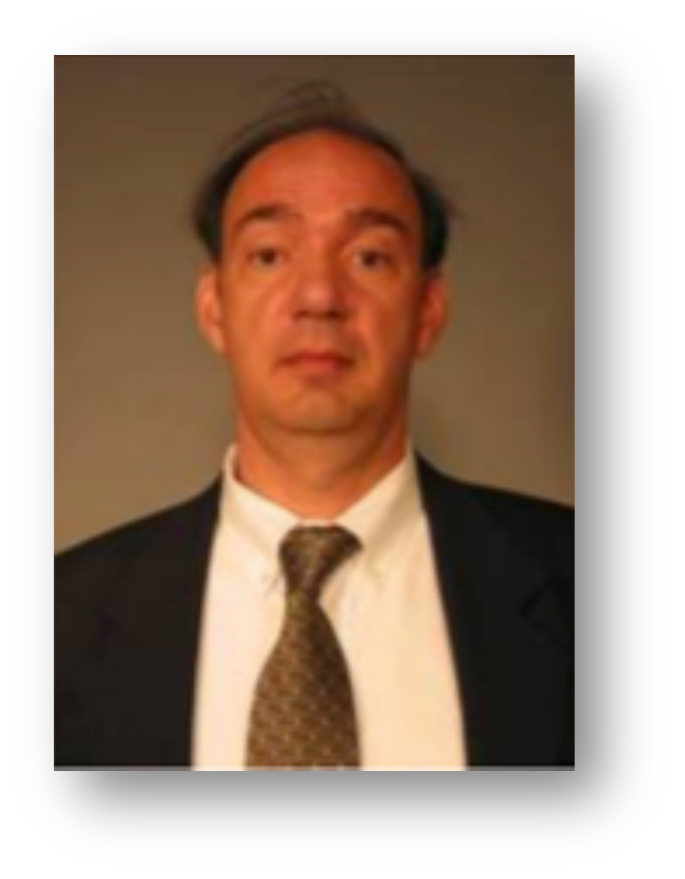}%
\includegraphics[width=0.7in]{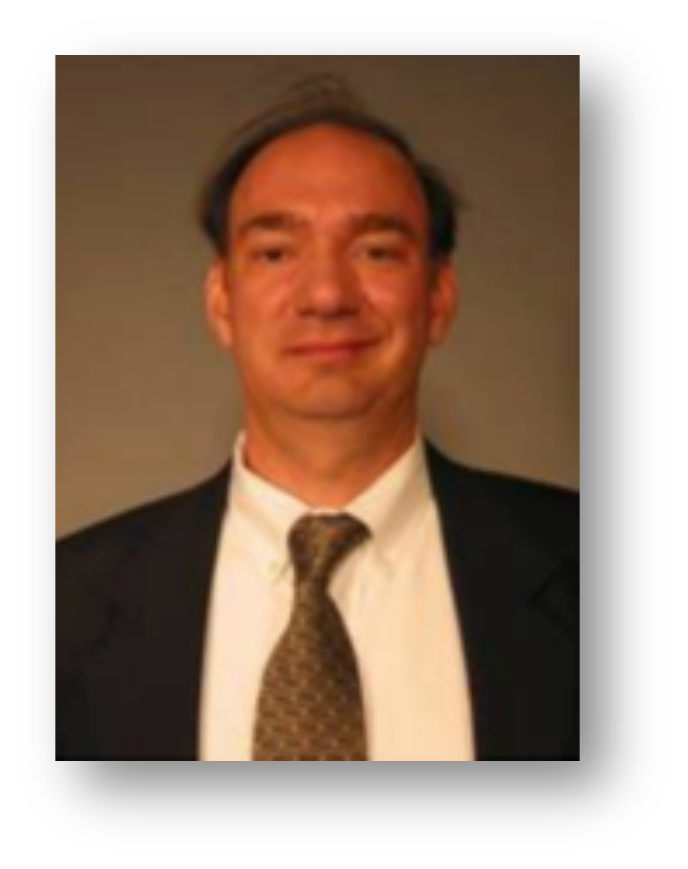}%
\includegraphics[width=0.7in]{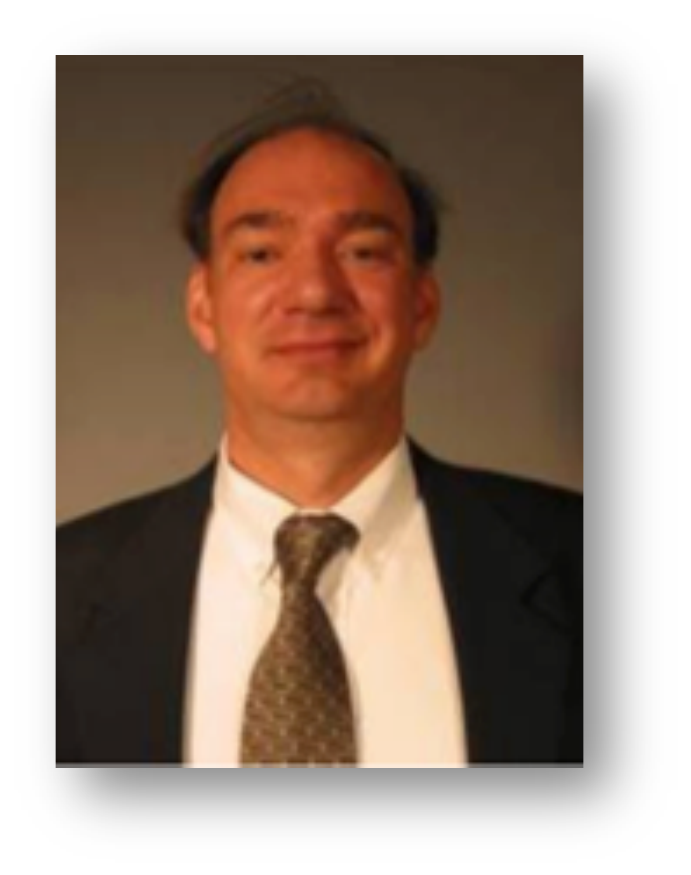}
\includegraphics[width=1.4in]{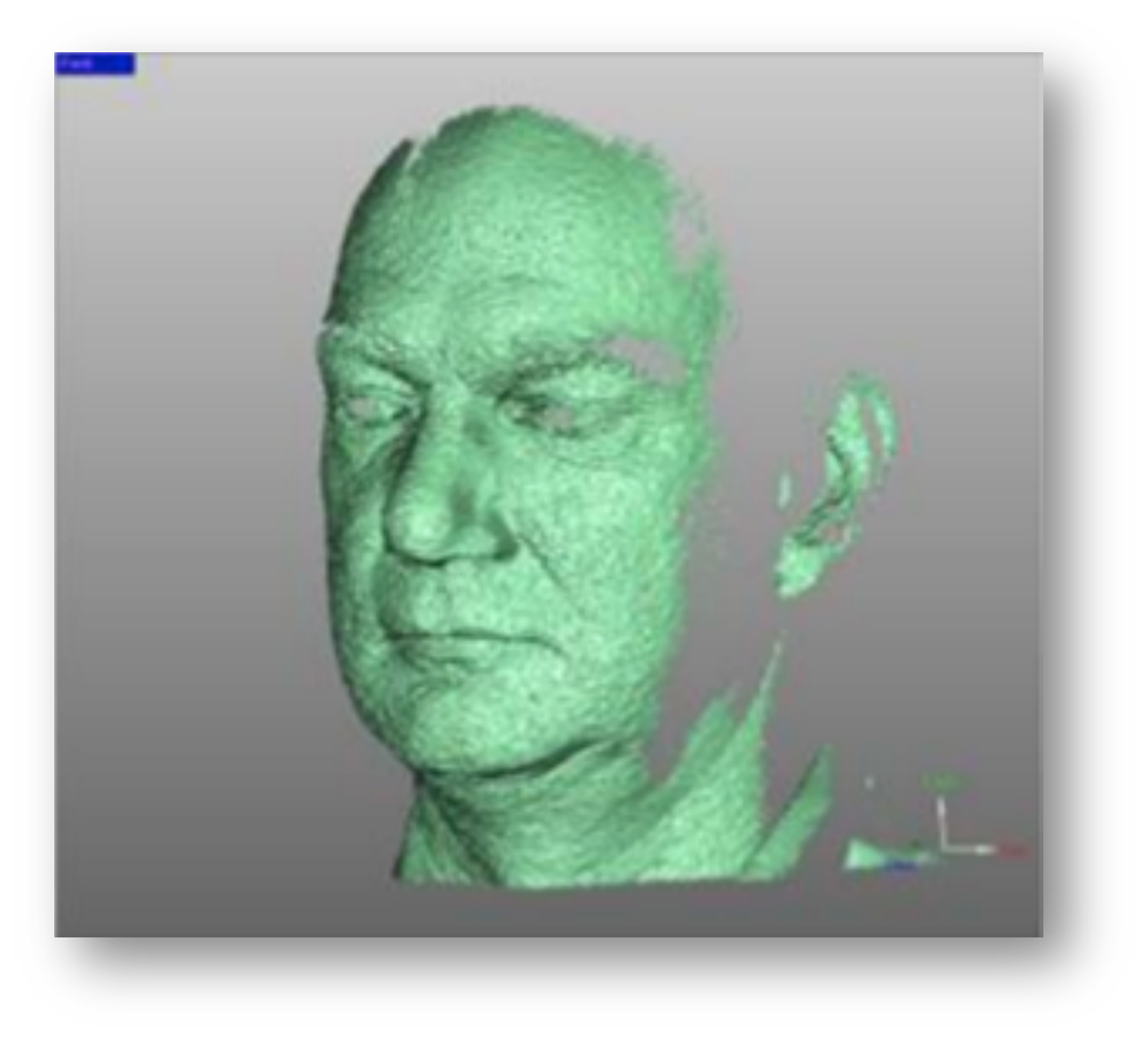}%
\includegraphics[width=1.4in]{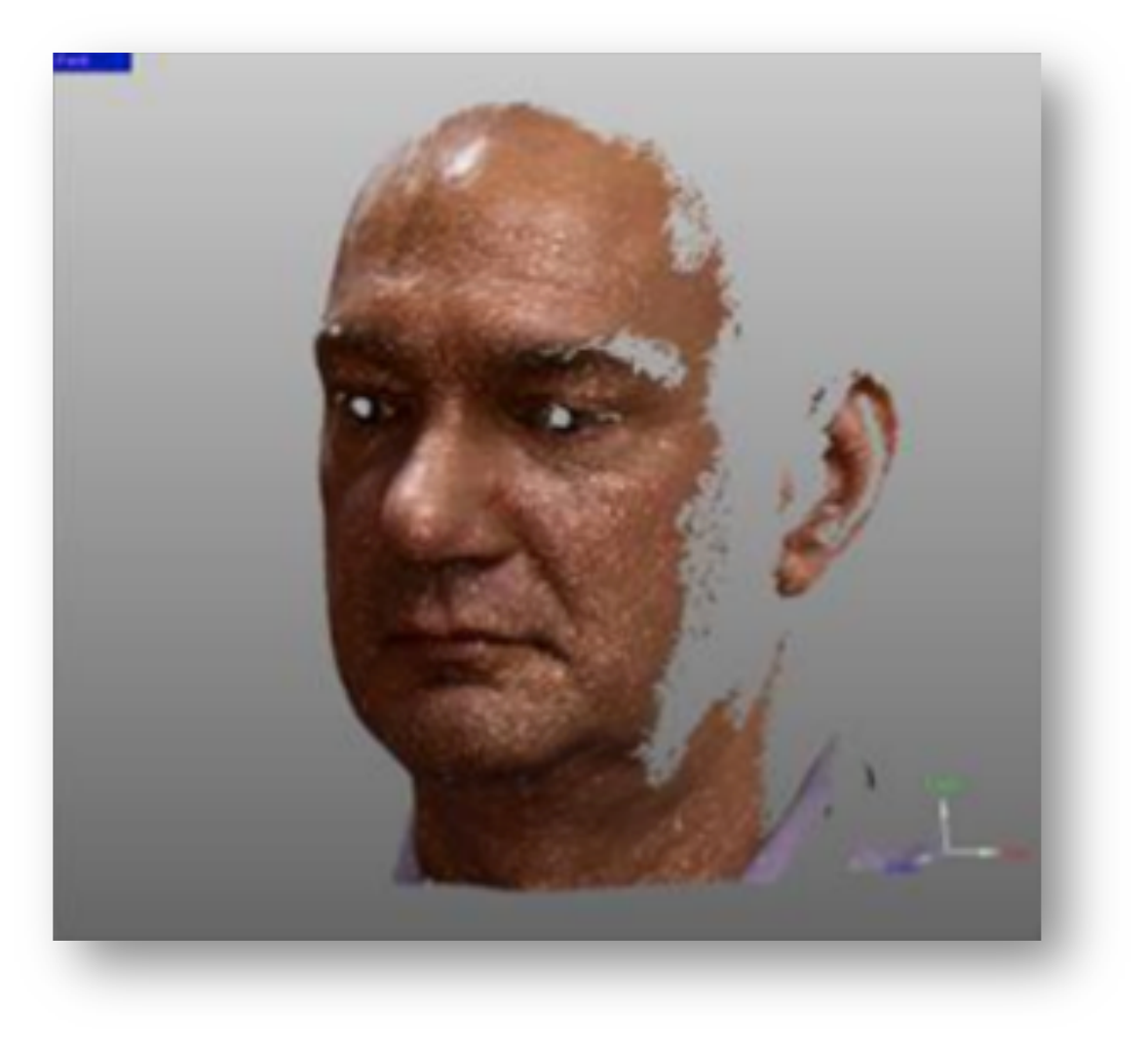}%
\end{center}
   \caption{2D images and 3D images from FRGC dataset.}
\label{fig:2D3D}
\end{figure}

The face Recognition Grand Challenge (FRGC) V2.0 dataset\footnote{Downloadable at http://www.nist.gov/itl/iad/ig/frgc.cfm} is widely used for 2D-3D face recognition. It consists of a total of 50,000 recordings spread evenly across 6,250 subjects. For each subject, there are 4 images taken in controlled light, 2 images taken under uncontrolled light and 1 3D image. The controlled images were taken in a studio setting while uncontrolled images were taken in changing illumination conditions. The 3D images were taken by a Minolta Vivid 900/910 series sensor, and both range and texture cues are included. An example from the FRGC V2.0 dataset is shown in Fig~\ref{fig:2D3D}.

UHDB11 \cite{Toderici:2014aa} is another popular dataset in 2D-3D face recognition. It consists of samples from 23 individuals, for each of which it has 2D high-resolution images spanning across six illumination conditions and 12 head-pose variations (72 variations in total), and a textured 3D facial mesh models. Each capture consists of both 2D images captured using a Canon DSLR camera and a 3D mesh captured by 3dMD 2-pod optical 3D system.

\subsection{Projection based approaches}

Yang et al.~\cite{yang2008cca2d3d} used CCA to correspond the 2D and 3D face modalities and deal with their heterogeneous dimensionality. Once projected into a common space, NN matching with Cosine distance is applied. To deal with the 2D-3D mapping being more complicated than a single linear transform, the CCA mapping is learned per-patch, and the matching scores fused at decision level.

Huang et al.~\cite{Liris-3912} presented a scheme to improve results by fusing 2D and 3D matching. 2D LBP features are extracted from both the 2D image and the 2D projection of the 3D image; and then compared with Chi-squared distance. Meanwhile LBP features are also extracted from both the 2D face and 3D range image. These are mapped into a common space using CCA and compared with cosine distance. The two scores are fused at decision level, and the desired result of 2D-3D matching outperforming 2D-2D matching is demonstrated.

To further improve recognition performance, Huang et al.~\cite{huang2010_2d3d} proposed a 2D-3D face recognition approach with  two separate  stages:  First, for 2D-2D matching, Sparse Representation Classifier (SRC) is used; Second, CCA is exploited to learn the projections between 3D and 2D face images. The two scores are again fused synergistically.

Petrou et al.~\cite{5539995} introduced a 2D-3D face recognition method based on a novel bidirectional relighting algorithm. A subject-specific 3D annotated model is built by using its raw 3D data and 2D texture. With this model, 2D images are projected onto a normalized image space. The lighting from the probe image is then transferred to the gallery image for more robust matching.

Rama et al.~\cite{rama2006_2d3d} generated 180-degree cylindrical face images in the 3D enrollment phase. At test time, after dimensionally reduction by $P^2CA$, 2D face images are compared against all subwindows of the 180-degree enrollment image. The score of the best-matching subwindow is taken as the score of the probe image.

\subsection{Feature based approaches}

A biologically inspired feature, Oriented Gradient Maps (OGMs), is introduced by Huang et al. in \cite{huang2012_2d3d}. OGMs simulate the complex neurons response to gradients within a pre-defined neighborhood. They have the benefit of being able to describe local texture of 2D faces and local geometry of 3D faces simultaneously. Using this feature, they are able to improve on both the 2D-2D and 2D-3D components of their previous work \cite{Liris-3912,huang2010_2d3d}.

\subsection{Summary and Conclusions}

2D image based face recognition systems often fail in situations where facial depictions exhibit strong pose and illumination variations. Introducing 3D models instead naturally solves these problems since poses are fully encoded and illumination can be modeled. However, matching 3D models generally is more computational resource demanding and incurs relatively higher cost (labor and hardware) in data acquisition. 2D-3D matching is thus gaining increasing interest as a middle ground to obtain improved pose invariance, with cheaper and easier data acquisition at test time.  In this area studies can be broken down into those that have done some kind of explicit 3D reasoning about matching the 2D probe image to the 3D model \cite{5539995,rama2006_2d3d}, and others that have relied on discriminative features and learning a single cross-domain mapping such as CCA \cite{Liris-3912,huang2010_2d3d,huang2012_2d3d,yang2008cca2d3d}. The latter approaches are somewhat more straightforward, but to fully realize the potential pose-invariance benefits of 2D-3D matching, methods that explicitly reason about pose mapping of each test image are likely to be necessary.

\section{Matching low and high-resolution face images}\label{sec:LR_HR}

The ability to match low-resolution (LR) to high-resolution (HR) face images has clear importance in security, forensics an surveillance. Interestingly we know this should be possible, because humans can recognize low-resolution faces down to $16\times16$ pixels \cite{sinha2006humanRecognition}. In practice, face images with high-resolution such as mug-shots or passport photos need to be compared against low-resolution surveillance images captured at a distance by CCTV, PTZ and wearable cameras. In this case there is a dimension mismatch between the LR probe images and HR gallery images. Simple image processing upscaling the probe images, or down-scaling the HR images is a direct solution to this, but it is possible to do better. 

In matching across resolution, existing approaches can be categorized into synthesis based  and projection-based. Synthesis based approaches, attempt to transform LR into HR images for matching. Super-resolution \cite{yang2010sparse_superres,ouwerkerk2006supres_survey} is used to reconstruct a HR representation of LR probe image. Then matching can be performed with any state of the art homogeneous face recognition systems. In projection-based approaches, HR gallery images and LR probes are projected into a common space in which classification is  performed.

\subsection{Synthesis based approaches}

Hennings-Yeomans et al.~\cite{4587810} presented a simultaneous super-resolution and recognition ($S^2 R^2$) algorithm to match the low-resolution probe image to high-resolution gallery. Training this algorithm learns a super-resolution model with the simultaneous objective that the resulting images should be discriminative for identity. In followup work, they further improved the super-resolution prior and goodness of fit feature used for classification \cite{5413920}.  However these methods have high computational cost.

Zou et al.~\cite{5957296} propose a similarly inspired discriminative super resolution (DSR) approach. The relationship between the two modalities is learned in the training procedure. Then, test time procedure, the learned relationship is used to reconstruct the HR images. In order to boost the effectiveness of the reconstructed HR images, a new discriminative constraint that exploits identity information in the training data is introduced. With these, the  reconstructed HR images will be more discriminative for recognition.

Zou et al.~\cite{Zou2010} proposed a nonlinear  super resolution algorithm to tackle LR-HR face matching. The kernel trick is used to tractably learn a nonlinear mapping from low to high-resolution images. A discriminative regularization term is then included that requires the high-resolution reconstructions to be recognizable.

Jia et al.~\cite{jia2005ssr} presented a bayesian latent variable approach to LR-HR matching. Tensor analysis is exploited to perform simultaneous super-resolution and recognition. This framework also has the advantage of simultaneously addressing other covariates such as view and lighting.

Jiang et al.~\cite{6467147} super-resolved LR probe images by Graph Discriminant Analysis on Multi-Manifold (GDAMM), before HR matching. GDAMM exploits manifold learning, with discriminative constraints to minimize within-class scatter and maximize across-class scatter. However to learn a good manifold multiple HR samples per person are required.

Huang et al.~\cite{5624630} proposed a nonlinear mapping based approach for LR-HR matching. First, CCA is employed to align the PCA  features of HR and LR face images. Then a nonlinear mapping is built with radial basis functions (RBF)s in this subspace. Matching is carried out by simple NN classifier.

Instead of  super-resolving a LR image for matching with HR images, Gunturk et al.~\cite{1203152} proposed an algorithm which constructs the information required by the recognition system directly in the low dimensional eigenface domain. This is more robust to noise and registration  than general pixel based super-resolution.  


\subsection{Projection-based approaches}

Li et al.~\cite{li2010lr} proposed a method that projects face images with different resolutions into a common feature space for classification. Coupled mappings that minimize the difference between the correspondences (i.e., low-resolution and its corresponding high-resolution image) are learned. The online phase of this algorithm is a simple linear transformation, so it is more efficient than many alternatives that perform explicit synthesis/super-resolution.  

Zhou et al.~\cite{6117595} proposed an approach named Simultaneous Discriminant Analysis (SDA). In this method, LR and HR images are projected into a common subspace by the mappings learned respectively by SDA. The mapping is designed to preserve the most discriminative information. Conventional classification methods can then be applied in the common space.

Wang et al.~\cite{Wang2013} present a projection-based approach called kernel coupled cross-regression (KCCR) for matching LR face images to HR ones. In this method, the relationship between LR and HR is described in a low dimensional embedding by a coupled mappings model and graph embedding analysis. The kernel trick is applied to make this embedding non-linear. They realize the framework with spectral regression to improve computational efficiency and generalization.

Sharma and Jacobs's cross-modality model \cite{sharma2011pls}  discussed previously can also be used for LR-HR matching. PLS is used to linearly map images of LR and HR to a common subspace.  The matching results show that PLS can be used to obtain state-of-the-art face recognition performance in matching LR to HR face images.

Multidimensional Scaling (MDS) is used by Biswas et al.~\cite{6112780} to simultaneously embed LR and HR images in a common space. In this common space, the distance between LR and HR approximates the distance between corresponding HR images.


Shekhar et al.~\cite{shekhar2011lrRec} proposed an algorithm to address low-high resolution face recognition, while maintaining illumination invariance required for practical problems. HR training images are relighted and downsampled, and LR sparse coding dictionaries are learned for each person. At test time LR images are classified by their reconstruction error using each specific dictionary.

Ren et al.~\cite{ren2012tip} tackle the low-high resolution face recognition by coupled kernel embedding (CKE). With CKE, they non-linearly map face images of both resolutions into an infinite dimensional Hilbert space where neighborhoods are preserved. Recognition is carried out in the new space.

Siena et al.~\cite{siena2013btas} introduced a Maximum-Margin Coupled Mappings (MMCM) approach for low-high resolution face recognition. A Maximum-margin strategy is used to learn the projections which maps  LR and HR data to a common space where there is the maximum margin of separation between pairs of cross-domain data from different classes. 

Ren et al.~\cite{ren2011mkl} addressed discriminative subspace learning in LR-HR recognition from the angle of evaluating and combining multiple encodings of each image type.  Different image descriptors including RsL2, LBP, Gradientface and IMED are considered and a multiple kernel learning strategy used to learn a good projection with a weighted combination of them.

In \cite{Li2009ACML}, Li et al. generalize CCA to use discriminative information in learning a low dimensional subspace for LR-HR image recognition. This is an closed-form optimization that is more efficient that super-resolution first strategies, while being applicable to other types of `degraded' images besides LR, such as blur and occlusion.

Deng et al.~\cite{5659338} utilized color information to tackle LR face recognition as color cues are less variant to resolution change. They improved on \cite{li2010lr} to introduce a regularized coupled mapping to project both LR and HR face images into a common discriminative space.

\subsection{Summary and Conclusions}

Both high-resolution synthesis and sub-space projection methods have been successfully applied to LR-HR recognition. In both cases the key insight to improve performance has been to use discriminative information in the reconstruction/projection, so that the new representation is both accurate and discriminative for identity.  Interestingly, while this discriminative cue has been used relatively less frequently in SBFR, NIR and 3D matching, it has been used almost throughout in HR-LR matching.

\paragraph{LR Dataset realism} With few exceptions \cite{shekhar2011lrRec}, the vast majority of LR-HR studies \emph{simulate} LR data by downsampling HR face images. Similarly to SBFR's focus on viewed-sketches, it is unclear that this is a realistic simulation of a practical LR-HR task. In practice, LR surveillance images are unavoidably captured with many other artefacts such as  lighting change, motion-blur, shadows, non-frontal alignment and so on \cite{hospedales2012spie,shekhar2011lrRec}. Thus existing systems are likely to under perform in practice. A benchmark dataset of realistic LR surveillance captures and associated HR mugshots would be advantageous to drive research. This may lead into integrating super-resolution and recognition with simultaneous de-blurring \cite{cho2007nonlocal_deblur,levin2011blinddeconv}, re-lighting \cite{shekhar2011lrRec} and pose alignment \cite{zhang2009faceRecAcrossPose}.

\section{Within-modality heterogeneity}
\label{within}

Independently of the sensing modality used, other covariates such as disguises and plastic surgery are also key factors that affect the performance of face recognition systems. These do not change the intrinsic quality or type of the images, but can still provide a strong image-space transformation between probe and gallery faces. Disguise is an interesting and challenging covariate of face recognition. It includes intentional or unintentional changes through which one can either hide his/her identity or appear to be someone else. Dhamecha et al.~\cite{6613019} have summarized the existing disguise detection and face recognition algorithms. In this survey, we rather focus on matching across plastic surgery variations.

With plastic surgery requiring reduced cost and time, its popularity has increased dramatically. 
It provides a significant covariate that seriously degrades the performance of conventional fare recognition systems, for example loosing 25-30\% rank 1 accuracy \cite{5492195}. Aside from its generally rising popularity, this result provides an incentive for individuals  to conceal their identity and evade recognition via plastic surgery. Hence it is of interest to develop HFR systems capable of recognition across pre and post-surgical images. 
\subsection{Database}
\begin{figure}[t]
\begin{center}
\includegraphics[height = 2.3in]{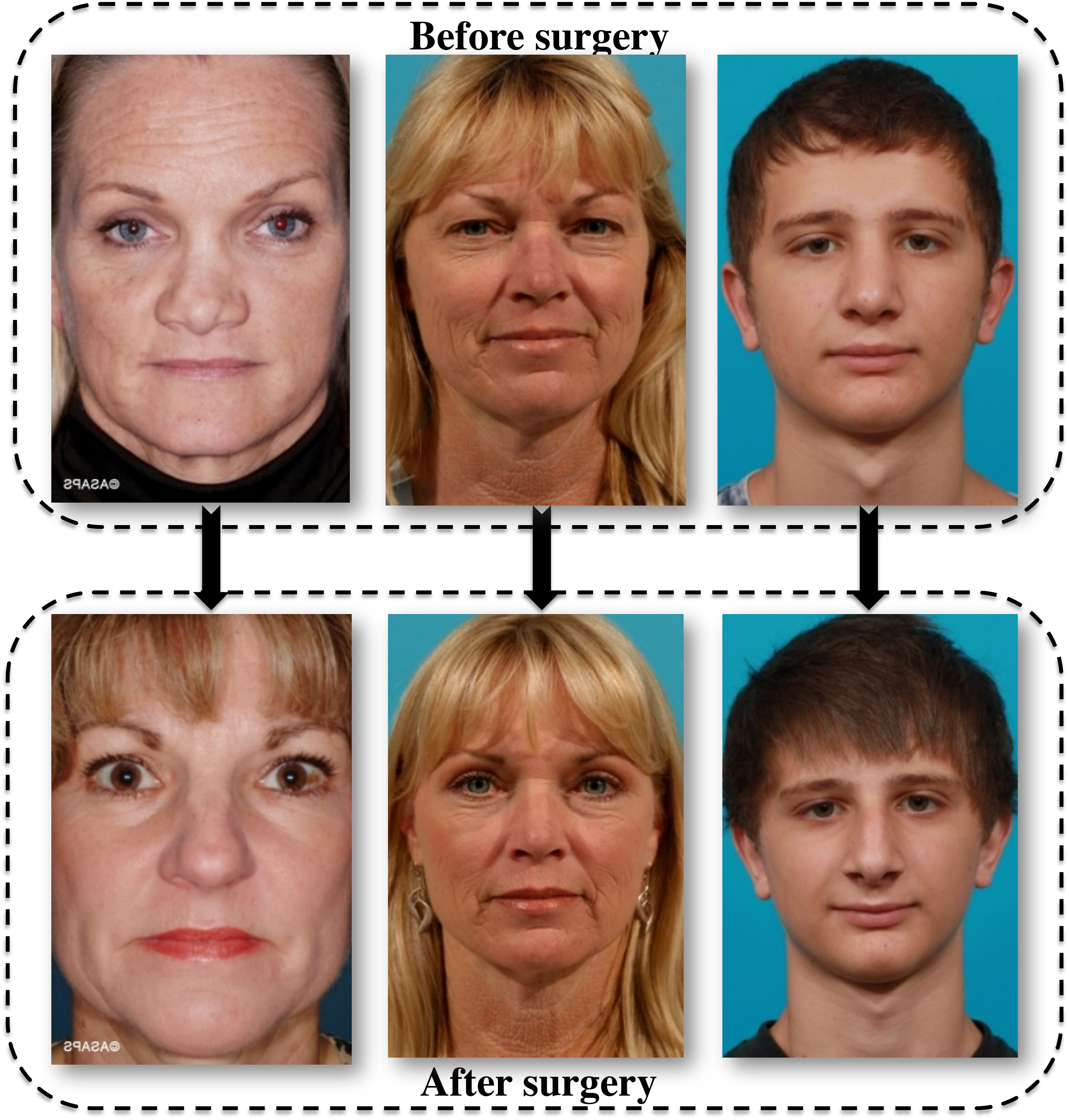}
\end{center}
   \caption{Pre and post-operative samples from the plastic surgery database.}
\label{fig:plastic}
\end{figure}

Singh et al.~\cite{5492195} provided a face database which encompasses 900 individuals who have plastic surgery. There are 900 subjects in the database corresponding to 1800 full frontal face images. A wide range of cases are included, such as nose surgery, as shown in Fig~(\ref{fig:plastic}), eyelid surgery, skin peeling, brow lift, and face lift. The details of images in the plastic surgery database are given in Tab~\ref{Tab:plasticdatabase}. 

\begin{table}[t]
\tbl{The details of plastic surgery database\label{Tab:plasticdatabase}}{%
\renewcommand{\arraystretch}{1.5}
\scalebox{0.8}{
\begin{tabular}{lc}
\whline
Plastic Surgery Procedure & Number of Individuals \\
\hline
\rowcolor{gray!20}
Dermabrasion& 32\\
Brow lift(Forehead surgery)& 60\\
\rowcolor{gray!20}
Otoplasty(Ear surgery) & 74\\
Blepharoplasty(Eyelid surgery) &105\\
\rowcolor{gray!20}
Rhinoplasty(Nose surgery) & 192\\
Others(Mentoplasty, Malar augmentation,&56\\
Craniofacial,Lip,augmentation,Fat,injection)  & \\
\rowcolor{gray!20}
Skin peeling(Skin resurfacing)& 73\\
Rhytidectomy(Face lift) & 308\\
\hline
\end{tabular}%
}
}
\end{table}

\subsection{Feature-based approaches}
Aggarwal et al.~\cite{6163008} locate facial components using active shape models, and then learn a sparse coding representation for each component. Components are matched across domains according to their sparse coding reconstruction errors. The overall face match is performed by sum fusion of the per-component scores.

Lakshmiprabha et al.~\cite{6416547} proposed a face recognition system invariant to plastic surgery using shape local binary texture (SLBT) feature in a two step cascade. In the first step, ASM is used to warp two images to be matched into alignment for global appearance based comparison -- thus partially addressing changes in face structure due to surgery. In the second step of the cascade per-component comparisons are made. However, this method requires manually annotated facial landmarks.

Bhatt et al.~\cite{6327663} developed an evolutionary granular algorithm to address the issues of automatic matching of face across plastic surgery variations. In this method, facial patches (granules) at multiple locations and resolutions are extracted, and two features (SIFT and EUCLBP) used to describe them. Evolutionary algorithms are used in order to select among granules, select the feature type for each, and determine their weighting in comparison using weighted Chi-square distance.

Liu et al.~\cite{Liu:2013aa}  proposed an ensemble of Gabor Patch classifiers via Rank-Order list Fusion (GPROF). Dividing face images into regular patches, Gabor features together with Fisher Linear Discriminant Analysis is exploited to generate a descriptor for each patch. The descriptors are then further transformed into a new invariant representation similar in inspiration to common representation \cite{Klare2010}: the rank ordering of their most similar gallery patches. Finally, the overall score fuses the result of each patch.

\subsection{Conclusions and discussion}

The key challenge of heterogeneity due to plastic surgery, is of course the intra-class variability introduced by the surgical process. Some previous cases of heterogeneity discussed in this survey such as sketch and NIR have more or less uniform and non-geometry distorting transformations (assuming good frontal poses). In contrast, surgical modifications can take a variety of forms including: similarly global but non structural/geometric modifications (e.g., skin resurfacing), global and structure/geometry distorting transforms (e.g., face lift), and highly localized transformations (e.g., nose surgery) \cite{5492195,6327663}. The multi-modality and non-uniformity of surgical transformations may explain why all of the studies so far have primarily been feature based approaches, rather than the synthesis and projection based approaches commonly seen in other HFR contexts. The prevalence of localized transformations also explains the heavier reliance in this area on component-based representations compared to other HFR settings. If a single facial component is modified, then the matching noise introduced is limited to the score of a single component.

\paragraph{Databases} An interesting issue is whether HFR systems for plastic surgery should be trained on non-surgery, or surgery databases. In the latter case, significantly more training data is likely to be available, but discriminatively trained models \cite{6327663} have then not been exposed to the variations which they will be tested on, and will thus under-perform. In the latter case the reverse is true, models will have been exposed to appropriate cross-modal variations at training time, but the amount of training data in HFR databases is less. These approaches were compared in \cite{5492195}, where training with 360 surgical pairs was reported to give better results than  on 900 non-surgical pairs. This issue in somewhat analogous to the previously mentioned question of whether to train on viewed or forensic sketches for forensic sketch recognition.

\section{Conclusion and Discussion}

As conventional within-modality face-recognition under controlled conditions approaches a solved problem, heterogeneous face recognition has grown in interest. This has occurred independently across a variety of covariates -- Sketch, NIR, LR, 3D and plastic surgery. In case there is a strong driving application factor in security/law-enforcement/forensics. We draw the following observations and conclusions:

\subsection{Common Themes}

\paragraph{Model types} Although the set of modality pairs considered has been extremely diverse (Sketch-Photo, VIS-NIR, HR-LR, 2D-3D), it is interesting that a few common themes emerge about how to tackle modality heterogeneity. Synthesis and subspace-projection have been applied in each case besides plastic surgery. Moreover, integrating the learned projection with a discriminative constraint that different identities should be separable, has been effectively exploited in a variety of ways. On the other hand, feature engineering approaches, while often highly effective have been limited to situations where the input-representation itself is not intrinsically heterogeneous (Sketch-Photo, and VIS-NIR).

\paragraph{Learning-based or Engineered} An important property differentiating cross-domain recognition systems is whether they require training data or not (and if so how much). Most feature-engineering based approaches have the advantage of requiring no training data, and thus not requiring a (possibly hard to obtain) dataset of annotated image pairs to be obtained before training for any particular application. On the other hand, synthesis and projection approaches (and some learning-based feature approaches), along with discriminatively trained matching strategies, can potentially perform better at the cost of requiring such a dataset.

\paragraph{Exploiting Face Structure} The methods reviewed in this survey varied in how much face-specific information is exploited; as opposed to generic cross-domain methods. Analytic and component-based representations of course exploit the specific face structure most heavily. Component-based methods are commonly used in recognition across plastic surgery. However, interestingly, the majority of methods reviewed do not exploit face-specific domain knowledge, relying on simple holistic or patch based representations with generally applicable synthesis/projection steps (e.g., CCA, PLS, sparse coding). Many methods do leverage the assumption of a fairly accurate and rigid correspondence in order use simple representations and mappings (such as patches with CCA). Going forward, this may be an issue in some circumstances like forensic sketch and realistic LR recognition where accurate alignment is impossible.

\paragraph{Dataset over-fitting} Recognition tasks in broader computer vision have recently been shown to suffer from over-fitting to entire datasets, as researchers engineer methods to maximize benchmark scores on insufficiently diverse datasets \cite{torralba2011dataset_bias}. Current HFR datasets, notably in Sketch, NIR and plastic surgery are also small and likely insufficiently diverse. As new larger and more diverse datasets are established, it will become clear whether existing methods do indeed generalize, and if the current top performers continue to be the most effective.

\subsection{Issues and Directions for Future Research}

\paragraph{Training data Volume} An issue for learning-based approaches is how much training data is required. Simple mappings to low-dimensional sub-spaces may require less data than more sophisticated non-linear mappings across modalities, although the latter are in principle more powerful. Current heterogeneous face datasets, for example in sketch \cite{wang2009faceSketchRec,bhatt2012memeticSketch,wang2009faceSketchRec,zhang2011informationTheoretic}, are much smaller than those used in homogeneous face recognition \cite{LFWTech} and broader computer vision \cite{deng2009imagenet}  problems. As larger heterogeneous datasets are collected in future, more sophisticated non-linear models may gain the edge.

\paragraph{Openness \& Components} Many studies make a contribution both to feature representation, and to some projection/synthesis/matching method. It is often hard to dis-entangle which part provides the benefit. It would be beneficial for the field if researchers: (i) always present their experiments breaking out feature and learning model contributions to the overall results, and (ii) released their features and learning methods, so that those who want to focus on one part can take best practice from the other without re-inventing the wheel.

\paragraph{Alignment} Unlike homogeneous face recognition which has moved onto recognition `in the wild' \cite{LFWTech}, heterogeneous recognition generally relies on accurately and manually aligned facial images. As a result, it is unclear how existing approaches will generalize to practical applications with inaccurate automatic alignment. Future work should address HFR methods that are robust enough to deal with residual alignment errors, or integrate alignment into the recognition process.

\paragraph{Side Information and Soft Biometrics} Side information and soft-biometrics have been used in a few studies \cite{5601735} to prune the search space to improve matching performance. The most obvious examples of this are filtering by gender or ethnicity. Where this information is provided as metadata, filtering to reduce the matching-space is trivial. Alternatively, such soft-biometric properties can be estimated directly from data, and then the estimates used to refine the search space. However, appropriate fusion methods then need to be developed to balance the contribution of the biometric cue versus the face-matching cue.

\paragraph{Facial Attributes} Related to soft-biometrics is the concept of facial attributes. Attribute-centric modelling has made huge impact on broader computer vision problems \cite{lampert2009zeroshot_dat}. They have successfully been applied to cross-domain modeling for person (rather than face) recognition \cite{layne2012attribreid}. Early analysis using manually annotated attributes highlighted their potential to help bridge the cross-modal gap by representing faces at a higher-level of abstraction \cite{klare2012charicature}. Recent studies \cite{sxoyBeyong} have begun to address fully automating the attribute extraction task for cross-domain recognition, as well as releasing facial attribute annotation datasets (both caricature and forensic sketch) to support research in this area. In combination with rapidly improving facial attribute extraction techniques  \cite{luo2013deepFacialAttr}, this is a promising avenue to bridge the cross-modal gap.

\paragraph{Computation Time} For automated surveillance, or search against realistically large mugshot datasets, we may need to attempt to recognise faces in milliseconds. Test-time computation is thus important, which may be an implication for models with sophisticated non-linear mappings across modalities; or in the LR-HR case, synthesis (super-resolution) methods that are often expensive.

\paragraph{Technical Methodologies} CCA, PLS, Sparse Coding and various generalizations thereof have been used extensively in the studies reviewed here. Some promising methodologies that have been under-exploited in HFR include metric learning and deep learning. 
Metric learning approaches have had great success in the related area of cross-view person recognition \cite{hirzer2012metricReid}. Early studies have shown the potential for HFR to improve the cross-domain and matching steps \cite{bhatt2012memeticSketch,6327663}. 

Deep learning  in contrast has transformed broader computer vision problems by learning significantly more effective feature representations \cite{krizhevsky2012imagenetDeepCNN}. Deep features may provide scope for bridging the cross-modal gap. For example, they have recently been applied for pose-invariant face-recognition \cite{zhu2013deepIdentityFace}. However, this requires  significantly larger scale heterogeneous datasets than is available for most of the HFR settings reviewed here.



\bibliographystyle{ACM-Reference-Format-Journals}
\bibliography{sketch_clean}




\end{document}